\documentclass[sigconf]{aamas} 

\usepackage{balance} 
\graphicspath{{images/}}
\usepackage{amsmath,amsfonts}
\usepackage{graphicx}
\usepackage{textcomp}
\usepackage{amsmath}
\usepackage{algorithm}
\usepackage{algpseudocode}
\usepackage{graphicx}
\usepackage{subcaption}
\usepackage{lipsum}
\usepackage{multicol}
\usepackage{hyperref}
\usepackage{xcolor}
\usepackage{wrapfig}
\usepackage{blindtext}
\usepackage{braket}

\setcopyright{ifaamas}
\acmConference[AAMAS '23]{Proc.\@ of the 22nd International Conference
on Autonomous Agents and Multiagent Systems (AAMAS 2023)}{May 29 -- June 2, 2023}
{London, United Kingdom}{A.~Ricci, W.~Yeoh, N.~Agmon, B.~An (eds.)}
\copyrightyear{2023}
\acmYear{2023}
\acmDOI{}
\acmPrice{}
\acmISBN{}

\acmSubmissionID{???}

\title[AAMAS-2023 Formatting Instructions]{AACHER: Assorted Actor-Critic Deep Reinforcement Learning with Hindsight Experience Replay}

\author{Adarsh Sehgal}
\affiliation{
  \institution{Advanced Robotics and Automation (ARA) Laboratory, Department of Computer Science and Engineering, University of Nevada}
  \city{Reno, Nevada}
  \country{USA}}
\email{asehgal@nevada.unr.edu}

\author{Muskan Sehgal}
\affiliation{
  \city{Reno, Nevada}
  \country{USA}}

\author{Hung Manh La}
\affiliation{
  \institution{Advanced Robotics and Automation (ARA) Laboratory, Department of Computer Science and Engineering, University of Nevada}
  \city{Reno, Nevada}
  \country{USA}}
\email{hla@unr.edu}

\begin{abstract}
Actor learning and critic learning are two components of the outstanding and mostly used Deep Deterministic Policy Gradient (DDPG) reinforcement learning method. Since actor and critic learning plays a significant role in the overall robot's learning, the performance of the DDPG approach is relatively sensitive and unstable as a result. We propose a multi-actor-critic DDPG for reliable actor-critic learning to further enhance the performance and stability of DDPG. This multi-actor-critic DDPG is then integrated with Hindsight Experience
Replay (HER) to form our new deep learning framework called AACHER. 
AACHER uses the average value of multiple actors or critics to substitute the single actor or critic in DDPG to increase resistance in the case when one actor or critic performs poorly. Numerous independent actors and critics can also gain knowledge from the environment more broadly. We implemented our proposed AACHER on goal-based environments: \textit{AuboReach}, \textit{FetchReach-v1}, \textit{FetchPush-v1}, \textit{FetchSlide-v1}, and \textit{FetchPickAndPlace-v1}. 
For our experiments, we used various instances of actor/critic combinations, among which A10C10 and A20C20 were the best-performing combinations. Overall results show that AACHER outperforms the traditional algorithm (DDPG+HER) in all of the actor/critic number combinations that are used for evaluation. When used on \textit{FetchPickAndPlace-v1}, the performance boost for A20C20 is as high as roughly 3.8 times the success rate in DDPG+HER.
\end{abstract}


\newcommand{\BibTeX}{\rm B\kern-.05em{\sc i\kern-.025em b}\kern-.08em\TeX}

\begin{document}


\pagestyle{fancy}
\fancyhead{}


\maketitle 


\section{Introduction}
A component of machine learning called Deep Learning employs hierarchical designs to extract high-level abstractions from data. It is a developing approach that has been employed in a variety of fields, like transfer learning \cite{tan2018survey}, the medical field \cite{bakator2018deep}, \cite{sehgal2022deep}, and many more \cite{sehgal2019lidar}, \cite{sehgal2019genetic}. Deep Reinforcement Learning (DRL) \cite{dong2020deep} techniques demonstrate robotics advancements. One of the constraints is the large number of interaction samples that are typically required for training in both simulated and real-world environments \cite{melnik2019tactile}. Autonomous robots have used Q-learning techniques for a range of tasks, and extensive study has been done in this area since many years ago, with some work focusing on continuous action spaces and others on discrete action spaces \cite{sehgal2019deep}. \cite{gu2016deep} Reinforcement learning (RL) \cite{sutton1998introduction} methods have been used to control a variety of robotic tasks, including locomotion \cite{kohl2004policy}, manipulation \cite{peters2008reinforcement}, \cite{theodorou2010reinforcement}, \cite{peters2010relative}, and autonomous vehicle control \cite{abbeel2006application}. One of the most successful applications of reinforcement learning is using robotic hands that passively adapt to environmental uncertainty. Soft body robots \cite{rus2015design} carry out straightforward tasks like gripping, which may help increase movement over soft materials. Similar applications of robotic hand manipulators can be also seen in \cite{sehgal2019genetic}, \cite{sehgal2019deep}, \cite{sehgal2022ga}, \cite{sehgal2022automatic}. 

There are three types of algorithms in reinforcement learning: actor-only, critic-only, and actor-critic method \cite{konda1999actor}. Recently, a novel deep reinforcement learning algorithm called deep deterministic policy gradient (DDPG) \cite{lillicrap2015continuous} has achieved good performance in many simulated continuous control problems. The deep deterministic policy gradient (DDPG) algorithm plays an important role in Actor-critic methods. The experience replay (ER) \cite{lin1992self} technique is a significant one in DDPG. One of the common experience replay techniques includes HER (Hindsight Experience Replay) \cite{andrychowicz2017hindsight} which enables sample-efficient learning from sparse and binary rewards, avoiding the need for complex reward engineering.   

As part of our research, we are contrasting our work with that of DDPG, using Hindsight Experience Replay (HER). The recent use of DDPG+HER was seen in \cite{sehgal2022automatic}. 
 
DDPG is made up of actor and critic learning. The performance of the algorithm is significantly influenced by both actors and critics. One of the studies \cite{wu2018multi} suggested the use of multi-critic learning (MCDDPG) to further enhance the stability and performance of DDPG and further used it with an extension of the experience replay mechanism known as Double Experience Replay (DER); where a more stable training process and preferable performance in practice were achieved.
In contrast with the previous works, our research makes use of \cite{baselines} to create a special algorithm that can function with multiple actors and critics in addition to HER. Actor-Critic Deep Reinforcement Learning with Hindsight Experience Replay, or AACHER, is the name we give to this system. In AACHER, the state-of-the-art DDPG algorithm is employed with multiple actor-critic schemes and their improvements are tested in various scenarios under varied circumstances. The algorithm is applied on five goal-based gym environments built on specially designed robotic manipulators: \textit{AuboReach}, \textit{FetchReach-v1}, \textit{FetchPush-v1}, \textit{FetchSlide-v1}, and \textit{FetchPickAndPlace-v1}. The entire algorithm is also examined to determine whether the technique is successful in enhancing the overall effectiveness of the learning process. The final results validate our assertion and offer convincing proof that the robot's overall learning process is enhanced by increasing the number of actors and critics.
Open source code is available at \textcolor{orange}{\href{https://github.com/aralab-unr/multi-actor-critic-ddpg-with-aubo}{https://github.com/aralab-unr/multi-actor-critic-ddpg-with-aubo}}.
 
The following is a list of our major contributions: 
\begin{itemize}
 \item Advanced DDPG combined with HER is used to create a novel algorithm known as AACHER (Actor-Critic Deep Reinforcement Learning with Hindsight Experience Replay).
 \item Advanced DDPG uses a variety of combinations of actors and critics to build independent instances for numerous actors, critics, or both.
 \item To further compute the loss and actions, we used the average of the actor and critic networks. Here, average refers to both the average of the actor networks and the average of the critic networks. The target networks, which apply the updated parameters, use the computed average for both actors and critics.
 \item We built Aubo-i5 custom environments (called \textit{AuboReach}) in both simulated and real-world settings to analyze the algorithm.
 \item Additionally, four of OpenAI's gym settings employ AACHER: \textit{FetchReach-v1}, \textit{FetchPush-v1}, \textit{FetchSlide-v1}, and \textit{FetchPickAnd Place-v1}.  
  \item Various combinations of actors and critics, using both simulated and real manipulation tasks, were used to examine the effectiveness of AACHER.
  \item The results of AACHER were compared to DDPG+HER.
  \item Our experiments show how AACHER outperforms DDPG+ HER which shows the effectiveness of having multiple actors and critics in DDPG.
\end{itemize}

Furthermore, this study is separated into several sections: Section II explains the background and motivation for Reinforcement Learning (RL), Deep Reinforcement Learning (DRL), Deep Deterministic Policy Gradients (DDPG), and Hindsight Experience Replay (HER). Section III explains our algorithm AACHER. Section IV explains our experiment settings, results, and discussions. Section V explains the conclusion and future work.

\section{Background \& Motivation}\label{chapter_two}

\subsection{Reinforcement Learning}
In typical reinforcement learning, a robot engages in discrete interactions with the environment to gather as many rewards from the environment as it can. A Markov Decision Process (MDP) \cite{puterman1990markov} can be used to model the issue as a tuple of $\Braket{\begin{array}{lcl}S, A, P, R, \gamma \end{array}}$ \cite{sutton2018reinforcement}, \cite{adam2011experience}. $S$ stands for the set of states, while $A$ stands for the set of actions. The transition probability function is $P:S \times A \times S \rightarrow [0, 1]$. $R: S \times A \rightarrow r \in \mathbb{R}$ is the reward function, and $\gamma$ is a scalar discount factor that corresponds to the same concept in Bellman Equation \cite{barron1989bellman}. Maximizing the expectation of discounted total rewards is the robot's aim: $R_t^\gamma = \sum_{i=t}^T \gamma^{i-t}r(s_i,a_i)$
by learning an ideal policy, where $t$ is the time step that ends at time $T$ and $r(s_t, a_t)$ represents the reward taking action $a_t$ in state $s_t$.


More specifically, the state-action value function with policy represents the long-term reward:

\begin{gather} 
    Q^\pi(s_t, a_t) = E[R_t^\gamma|s = s_t,a=a_t] \nonumber\\
    = E\left[\sum_{i=t}^T \gamma^{i-t}r(s_i,a_i)\right],
    \label{eq_rl_1}
\end{gather}

where the state-value function is the anticipated reward for a state if a robot follows a policy $\pi$:

\begin{gather} 
    V^\pi(s) = E[R_t^\gamma|S_t=s;\pi].
    \label{eq_rl_2}
\end{gather}

Actor and critic are defined as state-action value function $Q(s_t,a_t,\theta)$ and actor function $\mu(s_t,\omega)$ with parameters $\theta$ and $\omega$, respectively, in an actor-critic reinforcement learning \cite{grondman2012survey}. The actor picks up the rules following the advice of the critic, who provides an estimate of the total payment. It is common knowledge that the critic bases their understanding of the state-action value function on TD errors \cite{boyan2002technical}], by minimizing the squared TD error loss function:

\begin{gather} 
    L(\theta) = \left(r(s_t,a_t) + \gamma Q'(s_{t+1},a_{t+1},\theta) - Q(s_t, a_t, \theta)\right)^2.
    \label{eq_rl_3}
\end{gather}

However, when deep neural networks are employed to approximate the state-action value function, the critic tends to diverge, which results in the invalidity of an actor as well. Recent effective RL research on experience replay and target networks \cite{mnih2015human} is particularly useful in overcoming the challenge.
Since both of these apply to the DDPG approach, thorough descriptions of each are provided in the next subsection.

\subsection{DRL}

In-depth research has been done in this area since its start \cite{watkins1992q}, with some work focusing on continuous action spaces \cite{gaskett1999q, doya2000reinforcement, hasselt2007reinforcement, baird1994reinforcement} and others on discrete action spaces \cite{wei2017discrete}. Autonomous robots use Q-learning \cite{watkins1992q} techniques to perform a variety of tasks \cite{La_TCST2015}. Locomotion \cite{kohl2004policy, endo2008learning} and manipulation \cite{peters2010relative,kalakrishnan2011learning} have both benefited from the application of Reinforcement Learning (RL) \cite{sutton1998introduction}.


Off-policy and on-policy RL approaches are both available. On-policy methods, like SARSA learning \cite{sutton2018reinforcement}, aim to evaluate or enhance the policy upon which decisions are based. Off-policy methods \cite{munos2016safe} are beneficial for real robot systems. Examples include the Deep Deterministic Policy Gradient algorithm (DDPG) \cite{lillicrap2015continuous}, Proximal Policy Optimization \cite{schulman2017proximal}, Advantage Actor-Critic (A2C) \cite{mnih2016asynchronous}, and Normalized Advantage Function algorithm (NAF) \cite{gu2016continuous}. Robotic manipulators have also received a lot of attention \cite{deisenroth2011learning,7864333}. To achieve its objectives, some of this work relied on fuzzy wavelet networks \cite{lin2009h}, while others depended on neural networks \cite{miljkovic2013neural,duguleana2012obstacle}. A thorough survey of contemporary deep reinforcement learning (DRL) techniques for robot handling is given in \cite{nguyen2019review}. In theory, goal-conditioned reinforcement learning (RL) can teach a range of skills since it frames each activity in terms of the desired outcome \cite{chane2021goal}. 

A single robot \cite{pham2018,Pham_SSRR2018} as well as a multi-robot system \cite{La_CYBER2013,La_SMCA2015,pham2018cooperative,Dang_MFI2016,rahimi2018} have both undergone substantial RL training and instruction. Model-based and model-free learning algorithms have both been researched in the past. It has been demonstrated that model-free deep reinforcement learning algorithms are capable of learning a variety of tasks, from playing video games from images\cite{mnih2013playing} to mastering complicated locomotion techniques \cite{schulman2015trust}. Model-based learning algorithms generally rely on a model-based instructor to create deep network rules under realistic circumstances. Due to model bias, model-based algorithms typically perform worse asymptotically than model-free learners. DDPG is a prominent model-free DRL algorithm that is policy-based, which is why we chose it for our experiments. The performance of DDPG has recently benefited greatly from various productive RL studies, such as Experience replay \cite{lin1992self} and Target networks \cite{mnih2015human}. The robustness and sample efficacy of goal-achieving approaches is usually increased by using hindsight experience replay (HER), which is one method for experience replay. \cite{andrychowicz2017hindsight}. 
We are using DDPG in conjunction with Hindsight Experience Replay (HER) for our testing. \cite{nguyen2018deep} provides an overview of recent research on experience ranking's potential to accelerate DDPG + HER's learning rate. The subsequent subsection provides a thorough explanation of DDPG and HER.

\subsection{DDPG}
The Deep Deterministic Policy Gradients (DDPG) approach \cite{lillicrap2015continuous} integrates several approaches to address continuous control problems. The Deterministic Policy Gradients (DDPG) algorithm is an expansion of the DPG algorithm. The actor-critic technique, experience replay, target network, and deterministic policy gradient theorem make up the DDPG method, which was first proposed in \cite{silver2014deterministic}. The primary contribution is the demonstration of deterministic policy $\mu_\omega: S \rightarrow A$, which creates the precise action for the robot by providing the state rather than a probability distribution over all the actions. The performance objective according to the DDPG approach is:

\begin{gather} 
    J(\pi_\omega) = \int_S \rho^\pi(s) \int_A \pi_\omega(s,a)r(s,a)dads \nonumber\\
    = E_{s\sim\rho^\pi,a\sim\pi_\omega} \left[ r(s,a) \right],
    \label{eq_ddpg_1}
\end{gather}

where the state distribution is indicated by $\rho^\pi(s)$. In light of \cite{sutton2018reinforcement} and \cite{grondman2012survey}, the goal of the deterministic policy is:

\begin{gather} 
    J(\pi_\omega) = \int_S \rho^\pi(s)r(s,\mu_\omega(s))ds \nonumber\\
    = E_{s\sim\rho^\mu} [r(s,\mu_\omega(s))].
    \label{eq_ddpg_2}
\end{gather}

\begin{figure}[ht!]
\centering
  \begin{subfigure}[b]{\linewidth}
    \centering
    \includegraphics[width=5cm,height=4cm]{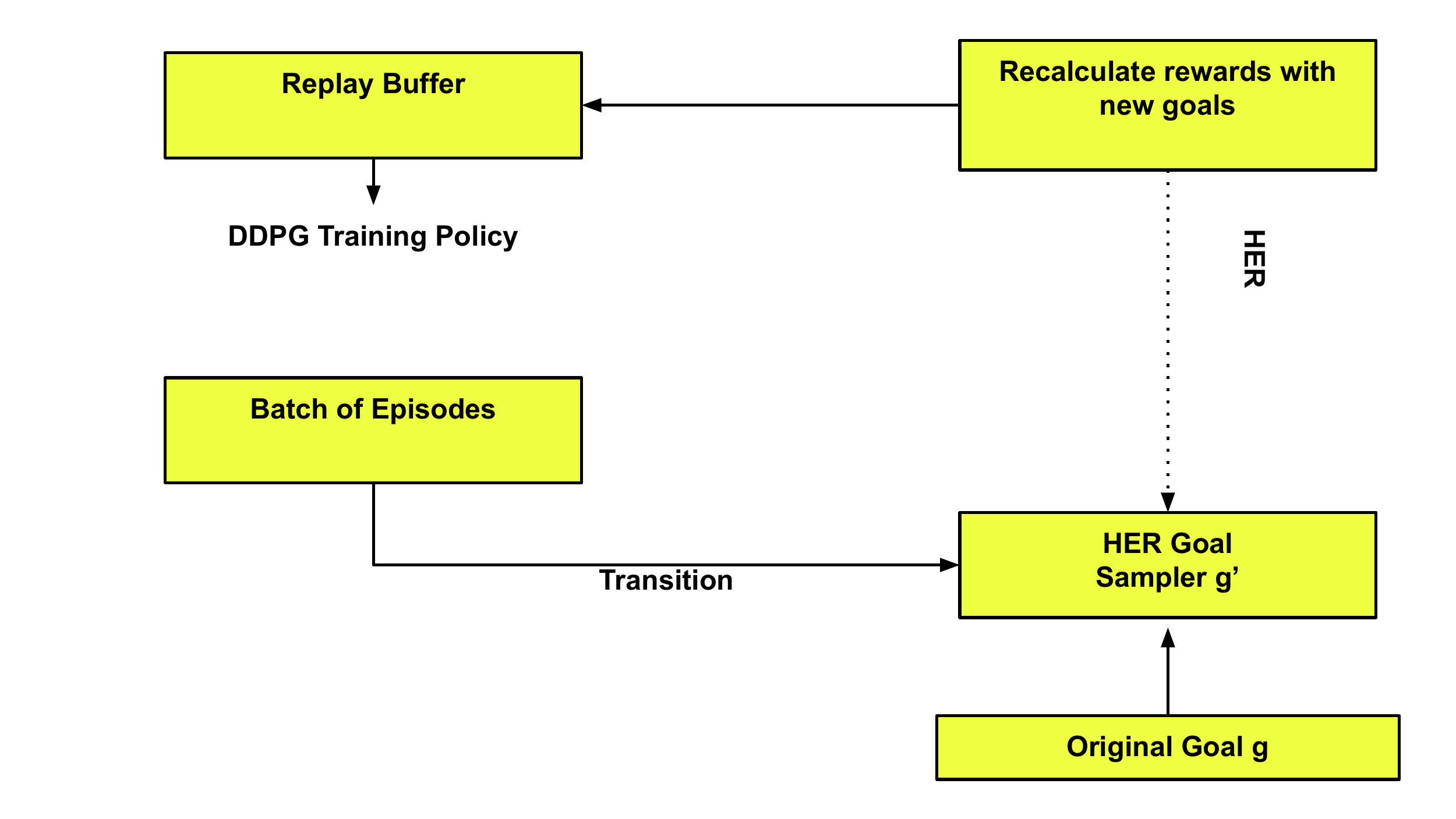}
  \end{subfigure}
  \caption{Explanation of DDPG+HER.}
  \label{fig:ddpg+her_diagram}
\end{figure}

The actor-network $\mu(s_t,\omega)$ and the state-action value (or critic) network $Q(s_t,a_t,\theta)$ are both intended to mimic the state-action value function and actor function, respectively, in the DDPG technique. The $\theta$ and $\omega$ are parameters of the neural networks. The experiences used for training are taken from the experience replay when the networks are updated. The 4-element tuple $(s_t,a_t,r_t,s_{t+1})$ is typically stored in a buffer that offers a batch of them for updating actor and critic networks. Only a small portion of the most recent experiences are retained since the newer experience will take the place of the older one when the buffer is filled. The target network, which is typically built as a replica of the main network, is also used to update the critic/actor network by providing a consistent target for the training process. In \ref{eq_rl_3}, the target network, designated as $Q^{tar}$, is used in place of $Q'(s_{t+1},a_t)$:

\begin{gather} 
    L_{tar}(\theta) = \left(r(s_t,a_t) + \gamma Q^{tar}(s_{t+1},a_{t+1},\theta^-) - Q(s_t,a_t,\theta) \right)^2,
    \label{eq_ddpg_3}
\end{gather}

where the parameter from a previous iteration is $\theta^-$.
To sustain the DDPG method's training and enable deep neural networks as well, the experience replay and target networks are equally crucial.

\subsection{HER}
Hindsight Experience Replay (HER) \cite{andrychowicz2017hindsight}  is a helpful method for Experience Replay that may be used in conjunction with any off-policy RL algorithm. It can be used whenever there are several goals. HER employs a sophisticated technique to teach robots more quickly in expansive state and action environments. By imitating human behavior, HER seeks to learn from failures. Even if the robot doesn't succeed in achieving the desired result, they learn from every experience. Whatever condition the robot meets is what HER regards as the amended aim. In ordinary experience replay, just the transition $(s_t||g,a_t,r_t,s_{t+1}||g)$ with the original goal $g$ is recorded. HER additionally stores the transition $(s_t||g',a_t,r'_t,s_{t+1}||g')$ to the changed goal $g'$. With really few incentives, HER functions wonderfully and is considered superior to shaped rewards in this aspect. Figure \ref{fig:ddpg+her_diagram} demonstrates how DDPG+HER functions. The HER algorithm selects new goals $g'$ at random from the batch of episodes \cite{nguyenhindsight}. These goals represent random states that a robot can attain throughout the episode. Then, rewards for these new additional goals will be calculated to produce new transitions linked to new goals, which will be saved in the replay buffer for policy training. The HER uses a future mechanism to select new goals, and the additional goals chosen are $k$ random states from the same episode as the transition being repeated.  

\section{Assorted Actor-Critic Deep
Reinforcement Learning with Hindsight Experience
Replay}\label{chapter_three}

\begin{algorithm}
\caption{AACHER}\label{algo:aacher}
\begin{algorithmic}[1]
    \State Create D Actor Neural networks
    \State Create P Critic Neural networks
    \State Initialize losses as an average of Actor and Critic neural networks
    \State Initialize replay buffer $R \gets \phi$
    \For{episode=1, M}
        \State Sample a goal $g$ and initial state $s_0$
        \For{t=0, T-1}
            \State Sample an action $a_t$ using behavioral policy generated by taking an average of policy neural networks
            \State Execute the action $a_t$ and observe a new state $s_{t+1}$
        \EndFor
        \For{t=0, T-1}
            \State $r_t:=r(s_t,a_t,g)$
            \State Store the transition $(s_t||g,a_t,r_t,s_{t+1}||g)$ in $R$
            \State Sample a set of additional goals for replay $G:=S$(\textbf{current episode})
            \For{$g'\in G$}
                \State $r':=r(s_t,a_t,g')$
                \State Store the transition $(s_t||g',a_t,r',s_{t+1}||g')$ in $R$ 
            \EndFor
        \EndFor
        \For{t=1,N}
            \State Sample a minibatch $B$ from the replay buffer $R$
            \State Perform one step of optimization using $A$ and minibatch $B$
        \EndFor
    \EndFor
\end{algorithmic}
\end{algorithm}

DDPG has demonstrated good performance in several areas, although there are still areas for performance enhancement and stability concerns \cite{peng2017sandpile}. The training of the DDPG technique is particularly sensitive to the efficiency of the actor/critic learning process because the actor's learning process largely depends on the critic. By using the DDPG method and an average actor/critic, it is possible to obtain an evident improvement in stability and performance. The details of the multi-actor/critic DDPG+HER approach are presented in this section (AACHER).

\begin{figure}[ht!]
\centering
  \begin{subfigure}[b]{\linewidth}
    \centering
    \includegraphics[width=8cm,height=5.5cm]{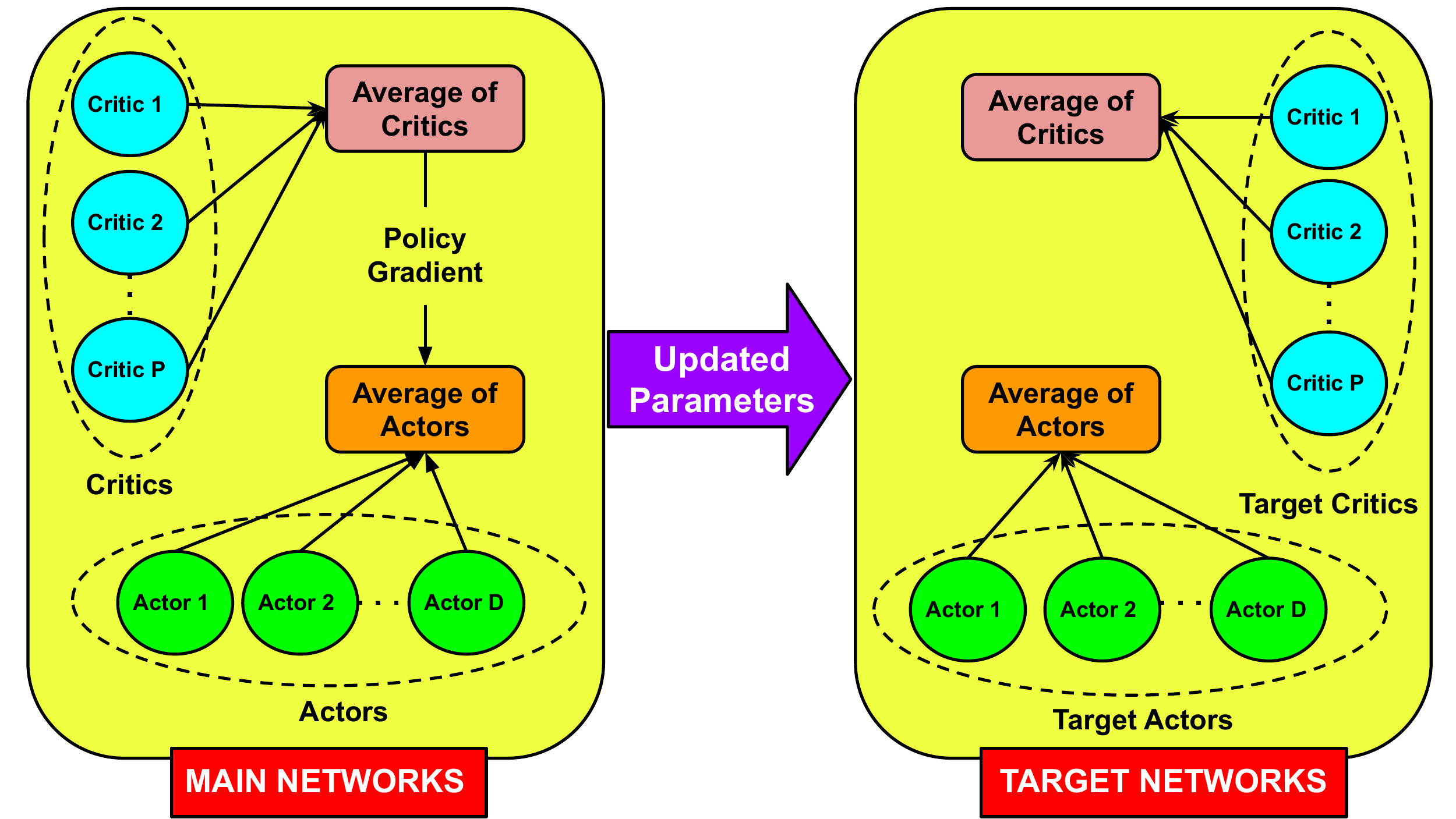}
  \end{subfigure}
  \caption{AACHER using multiple actors and critics.}
  \label{fig:aacher_diagram}
\end{figure}


Because AACHER employs $D$ actors and $P$ critics in its actor-critic architecture, the actor network is represented by the average of $D$ actor values, and the state-action value function is estimated by the average of $P$ critic values:

\begin{gather} 
    \mu_{avg}(s,\omega) = \frac{1}{D}\sum_{i=1}^{D}\mu_{i}(s,\omega_i),
    \nonumber\\
    Q_{avg}(s,a,\theta) = \frac{1}{P}\sum_{i=1}^{P}Q_i(s,a,\theta_i),
    \label{eq_aacher_avg}
\end{gather}

where $\omega_i \in \omega$ and $\theta_i \in \theta$ represent the i-th actor and critic parameters, respectively. In contrast to the actions/Q-values that were previously learned, the AACHER technique builds D/P independent actor/critic networks. As a result, when one actor or critic gives a terrible performance, the average of all performers or actors/critics will somewhat mitigate the negative impact. Additionally, numerous independent actors and critics can gain a broader understanding of the environment. The actor networks are made up of a parameterized group of policies that are typically updated by a policy gradient, while the critic networks are updated by TD errors.

Training of critics uses TD errors between the average of critics and target critics:

\begin{gather} 
    L_{avg}(\theta) = \left(r(s,a) + \gamma Q_{avg}^{tar}(s,a,\theta^{-}) - Q_{avg}(s, a, \theta)\right)^2.
    \label{eq_rl_3}
\end{gather}

\begin{figure}
\centering
  \begin{subfigure}[b]{\linewidth}
    \centering
    \includegraphics[width=0.24\linewidth,height=2.2cm]{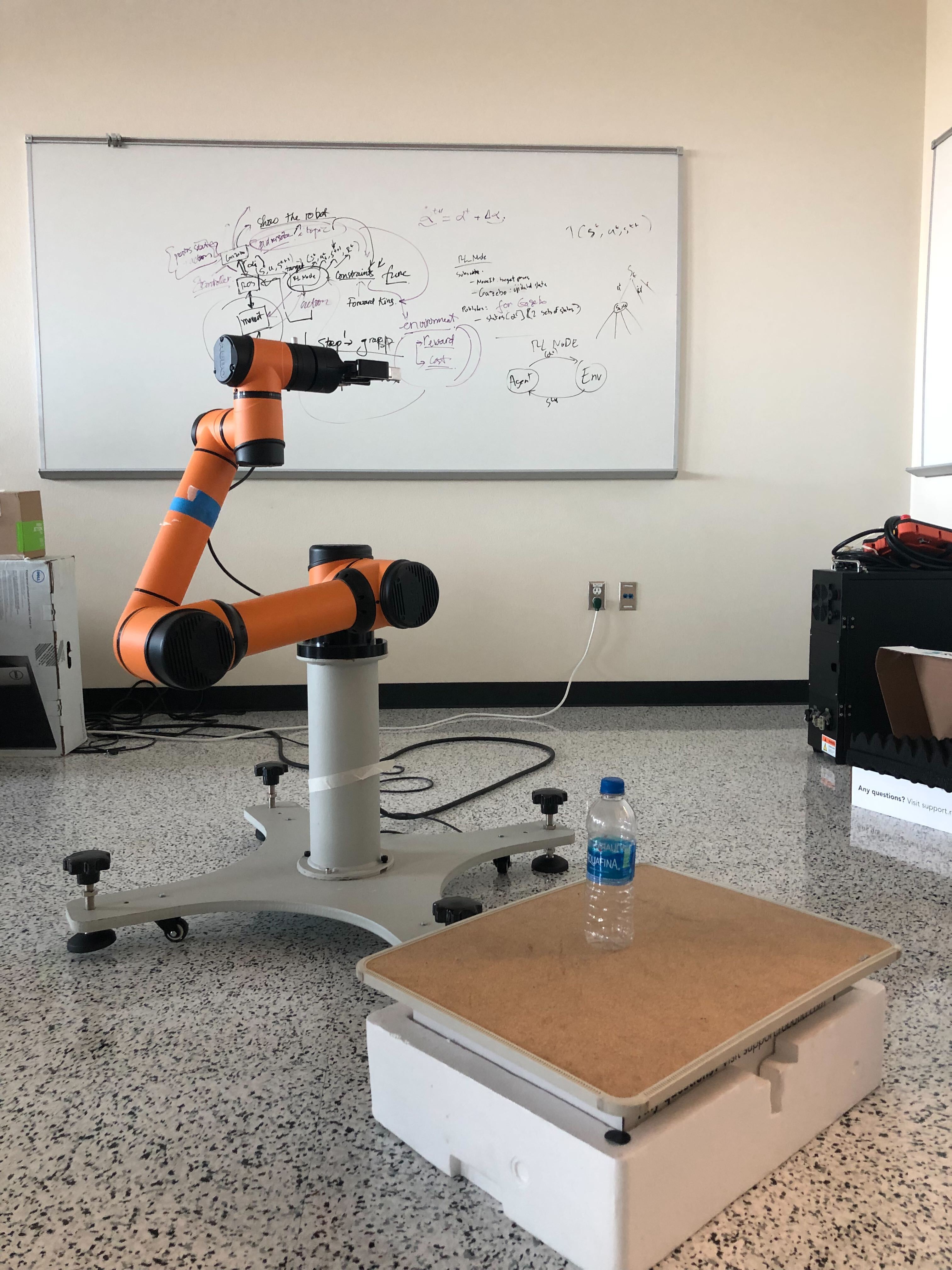}
    \includegraphics[width=0.24\linewidth,height=2.2cm]{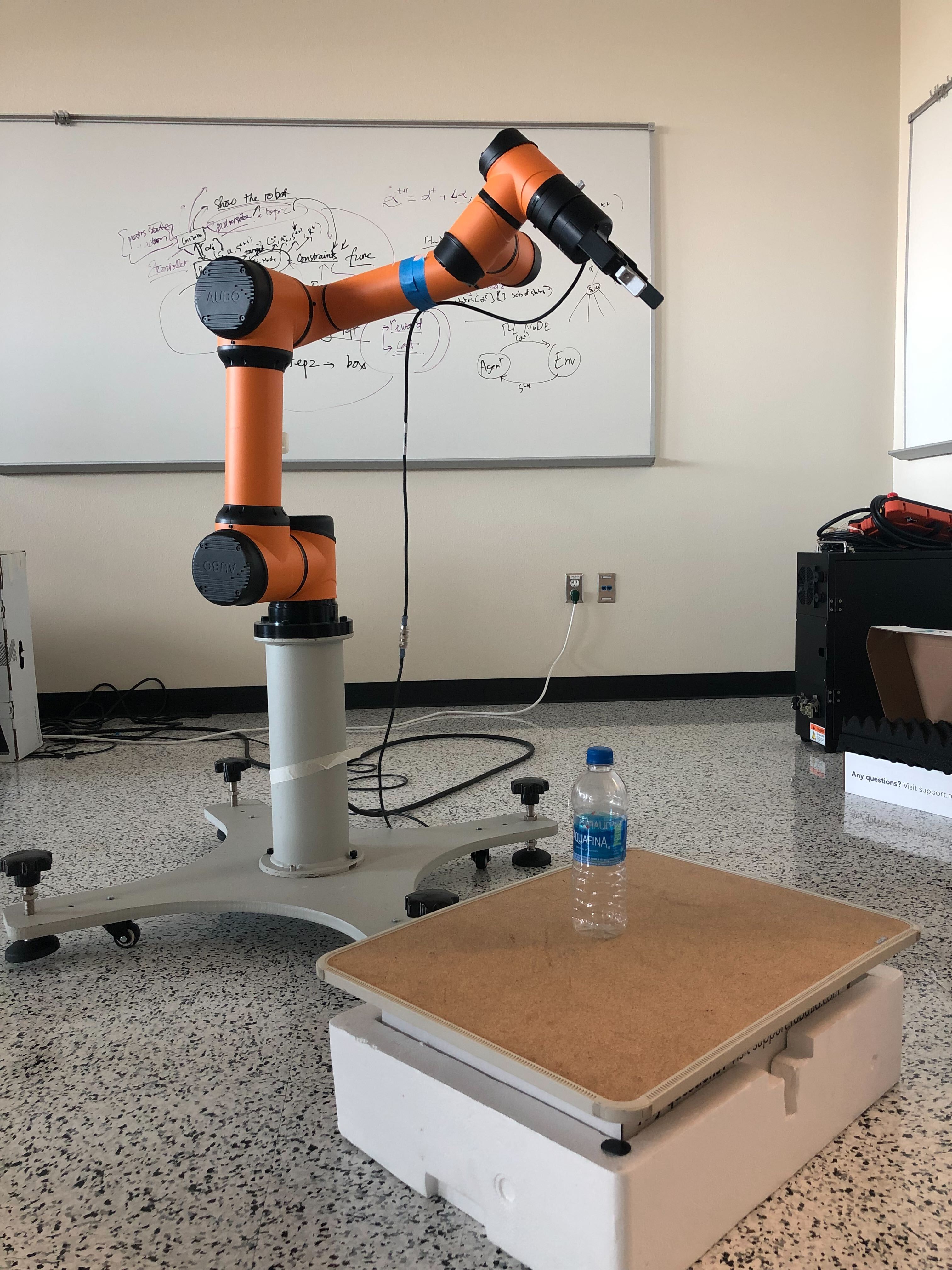}
    \includegraphics[width=0.24\linewidth,height=2.2cm]{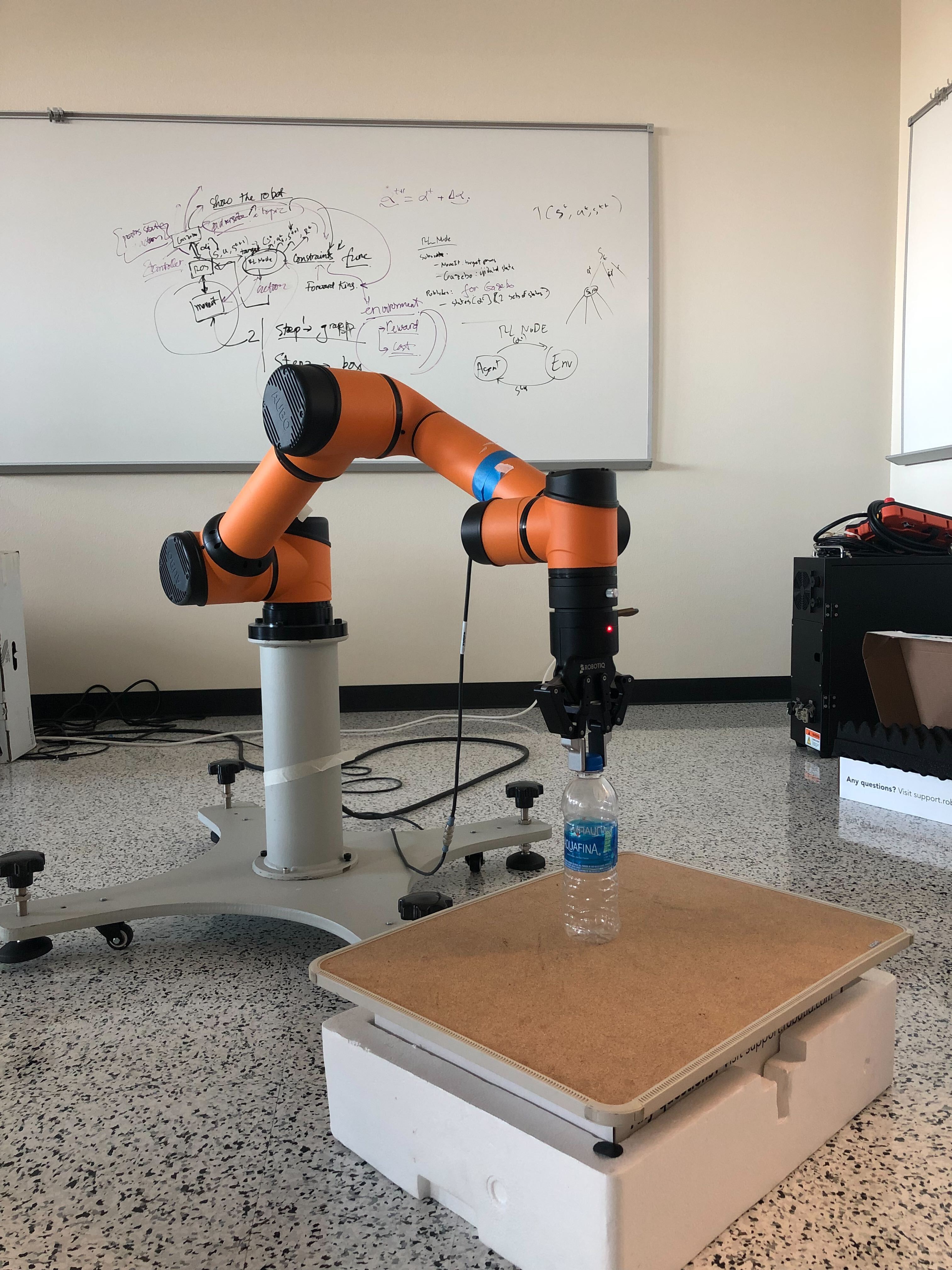}
    \includegraphics[width=0.24\linewidth,height=2.2cm]{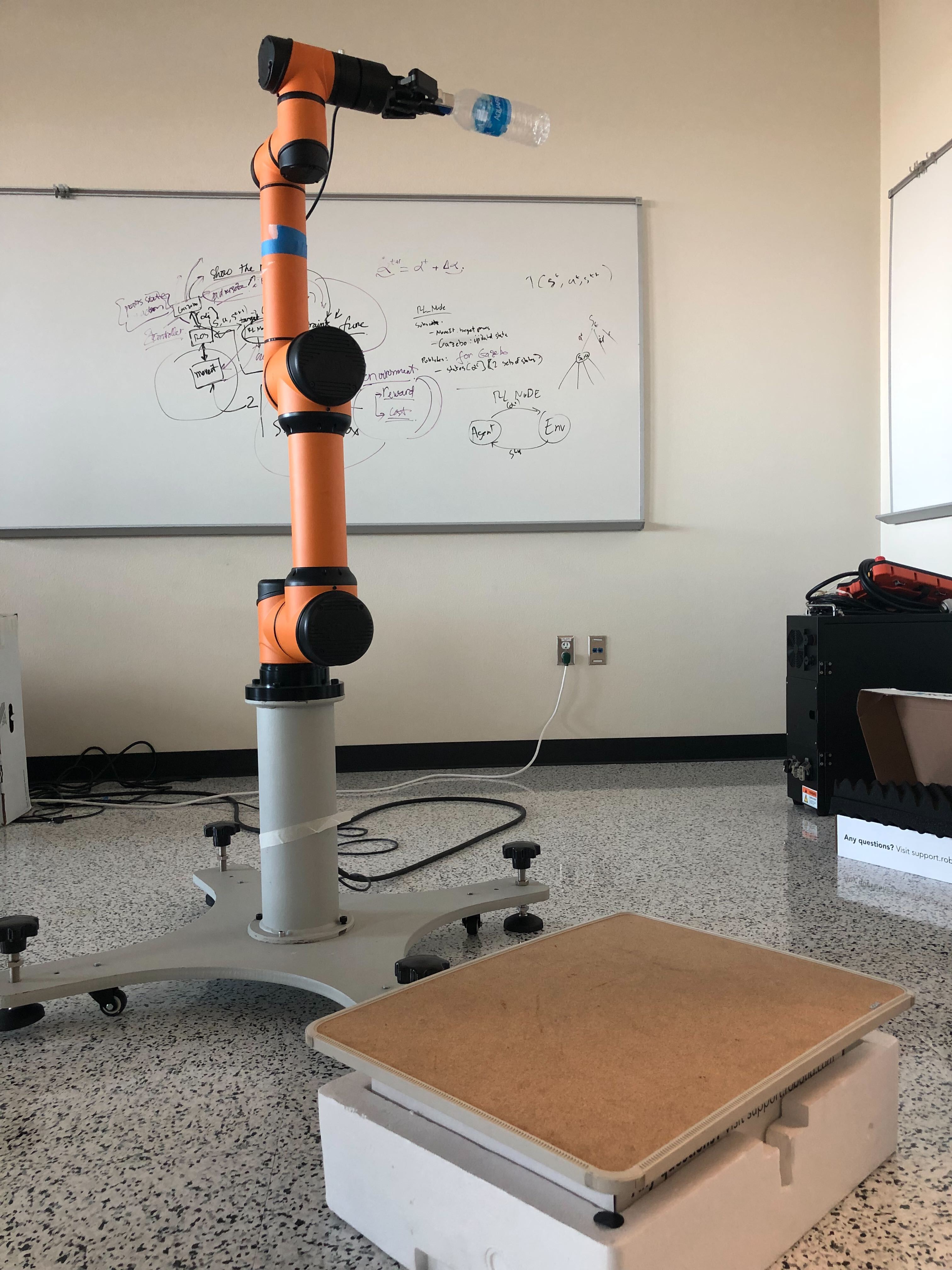}
  \end{subfigure}
  \caption{Using the most accurate policy learned via AACHER, the \textit{AuboReach} environment performs a task in a real experiment.}
  \label{fig:auboReachEnv}
\end{figure}

\begin{figure}
\centering
  \begin{subfigure}[b]{\linewidth}
    \centering
    \includegraphics[width=0.24\linewidth,height=2.2cm]{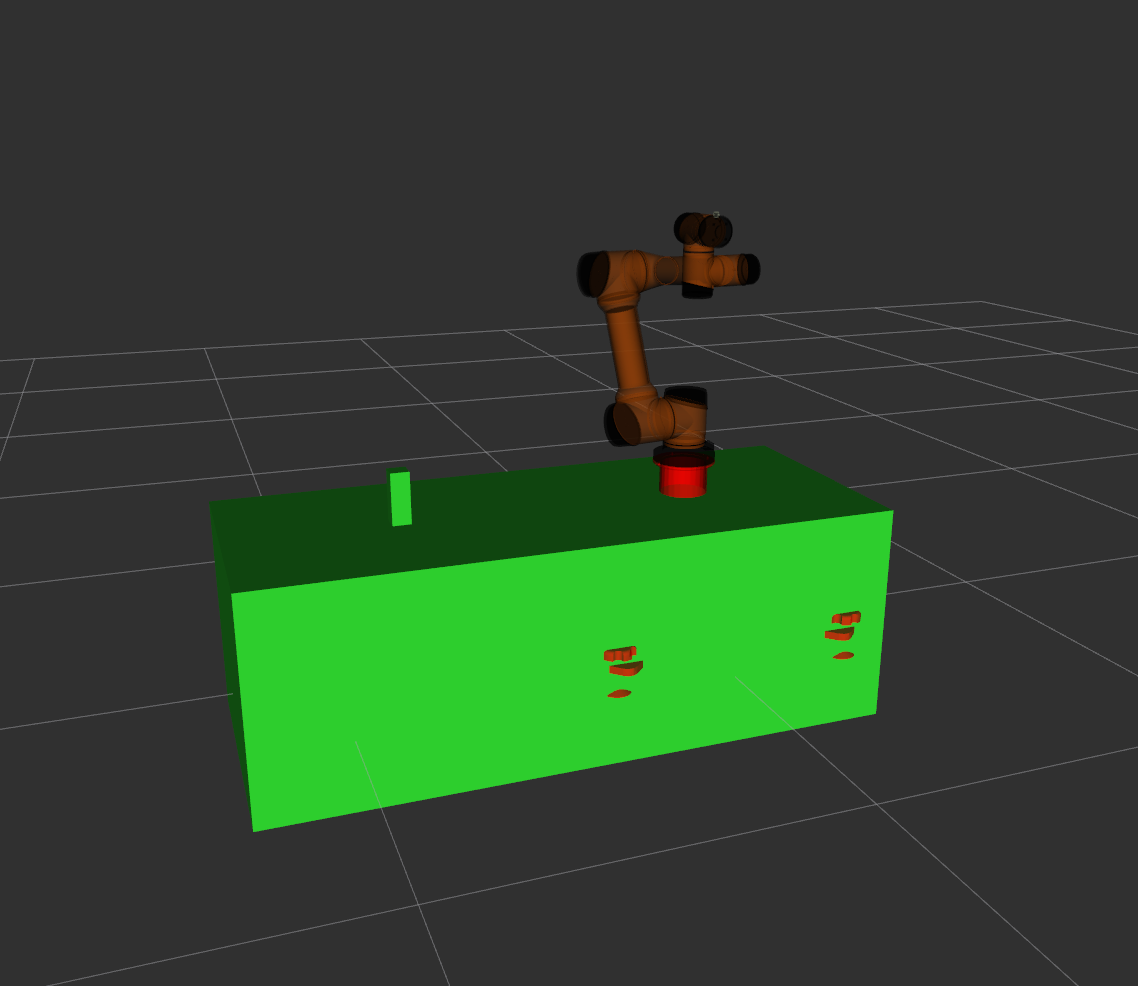}
    \includegraphics[width=0.24\linewidth,height=2.2cm]{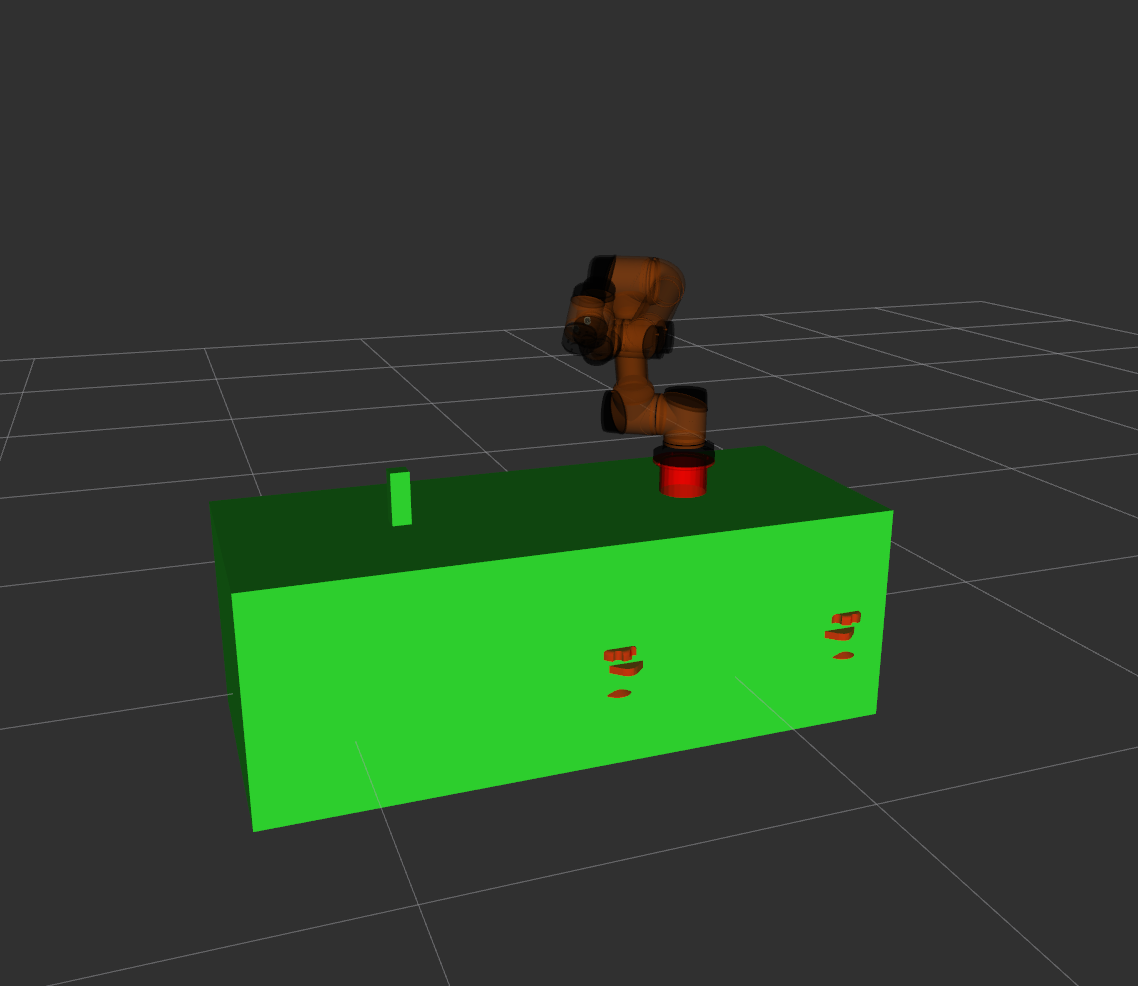}
    \includegraphics[width=0.24\linewidth,height=2.2cm]{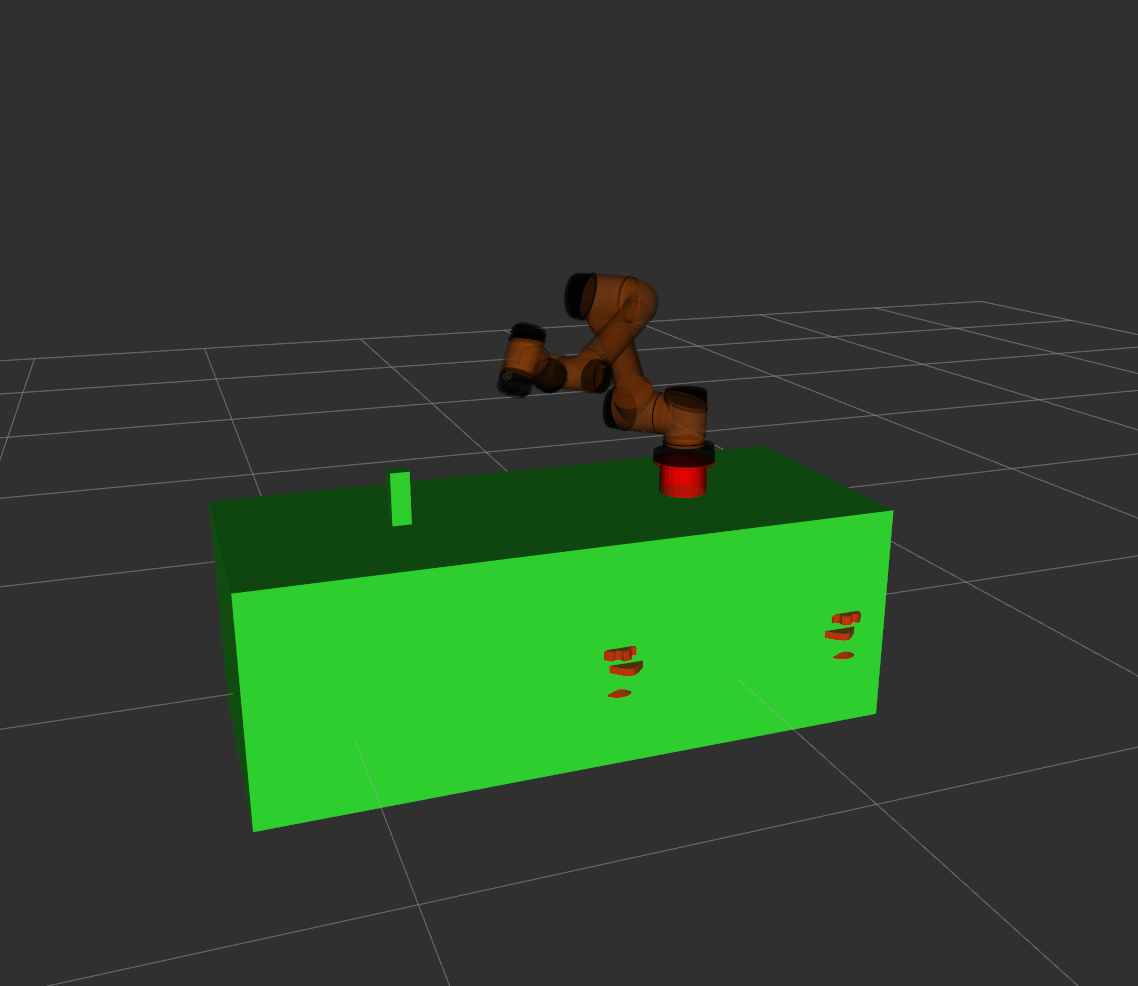}
    \includegraphics[width=0.24\linewidth,height=2.2cm]{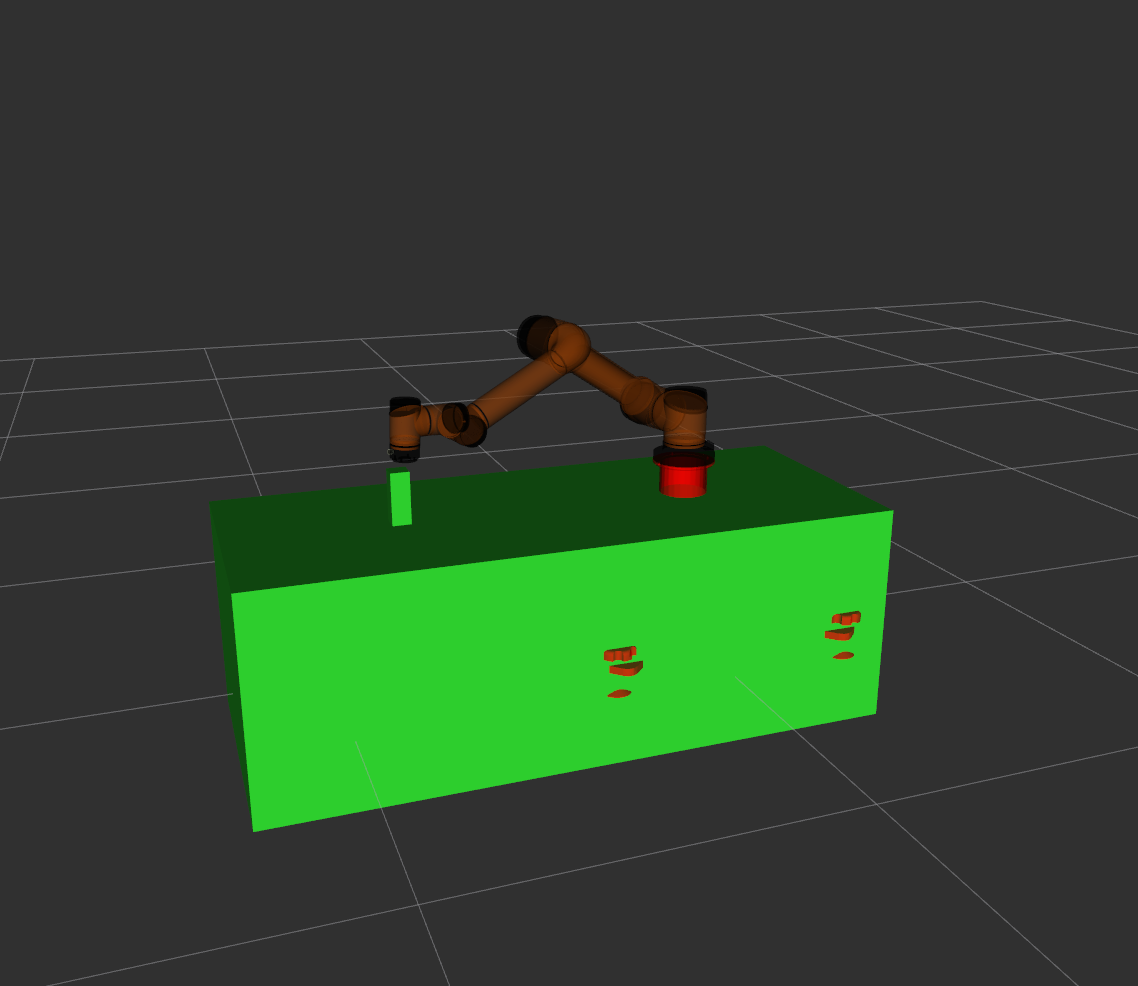}
  \end{subfigure}
  \caption{In a simulated experiment, the \textit{AuboReach} environment performs a task using the best policy learned via AACHER.}
  \label{fig:auboReachEnvSim}
\end{figure}

The AACHER approach is concretely described in Algorithm \ref{algo:aacher}, and the fundamental design of DDPG and HER remains unchanged. Here in the algorithm, line 15 states additional goals which are sampled from $g'\in G$. The ratio of the total number of elements in $g'$ and $G$ is controlled by the hyperparameter $k$. Here, $k$ random states from the same episode as the replayed transition will be used to choose additional goals. A visual illustration of AACHER is shown in Figure \ref{fig:aacher_diagram}. It displays average actors and critics across main and target networks. We devised a naming standard so that we could quickly identify the many tests we did with AACHER. 
The experiments are referred to by the acronym \textit{ADCP}, where A stands for an actor, D for the number of actors, C for the critic, and P for the number of critics. As an illustration, A2C3 denotes the usage of 2 actors and 3 critics in each main and target network, which are then averaged for training.

\section{Experiments}\label{chapter_four}

\begin{figure*}
\centering
  \begin{multicols}{3}
  \includegraphics[width=\linewidth,height=4cm]{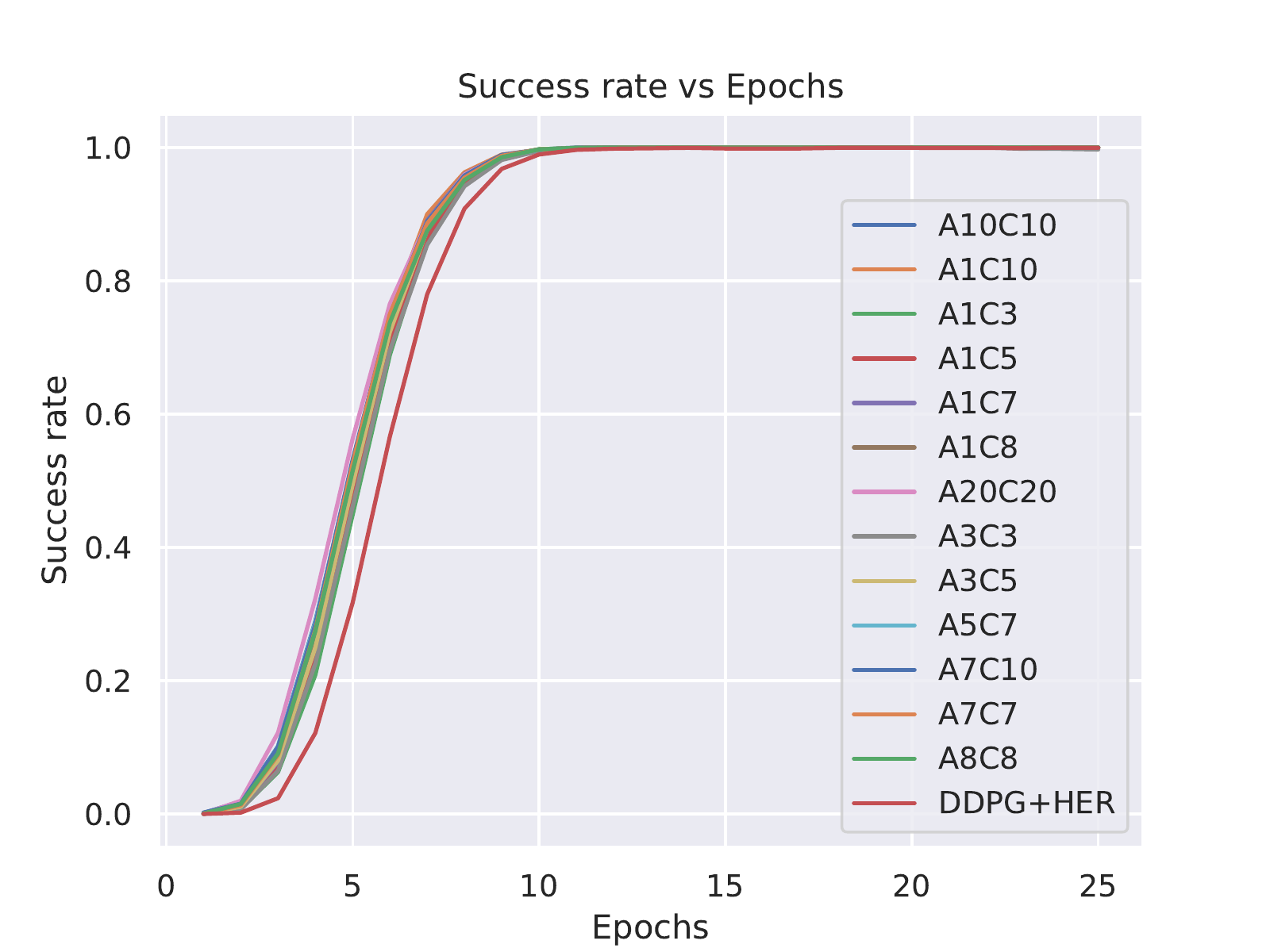}
  \subcaption{Success rate vs Epochs}
  \includegraphics[width=\linewidth,height=4cm]{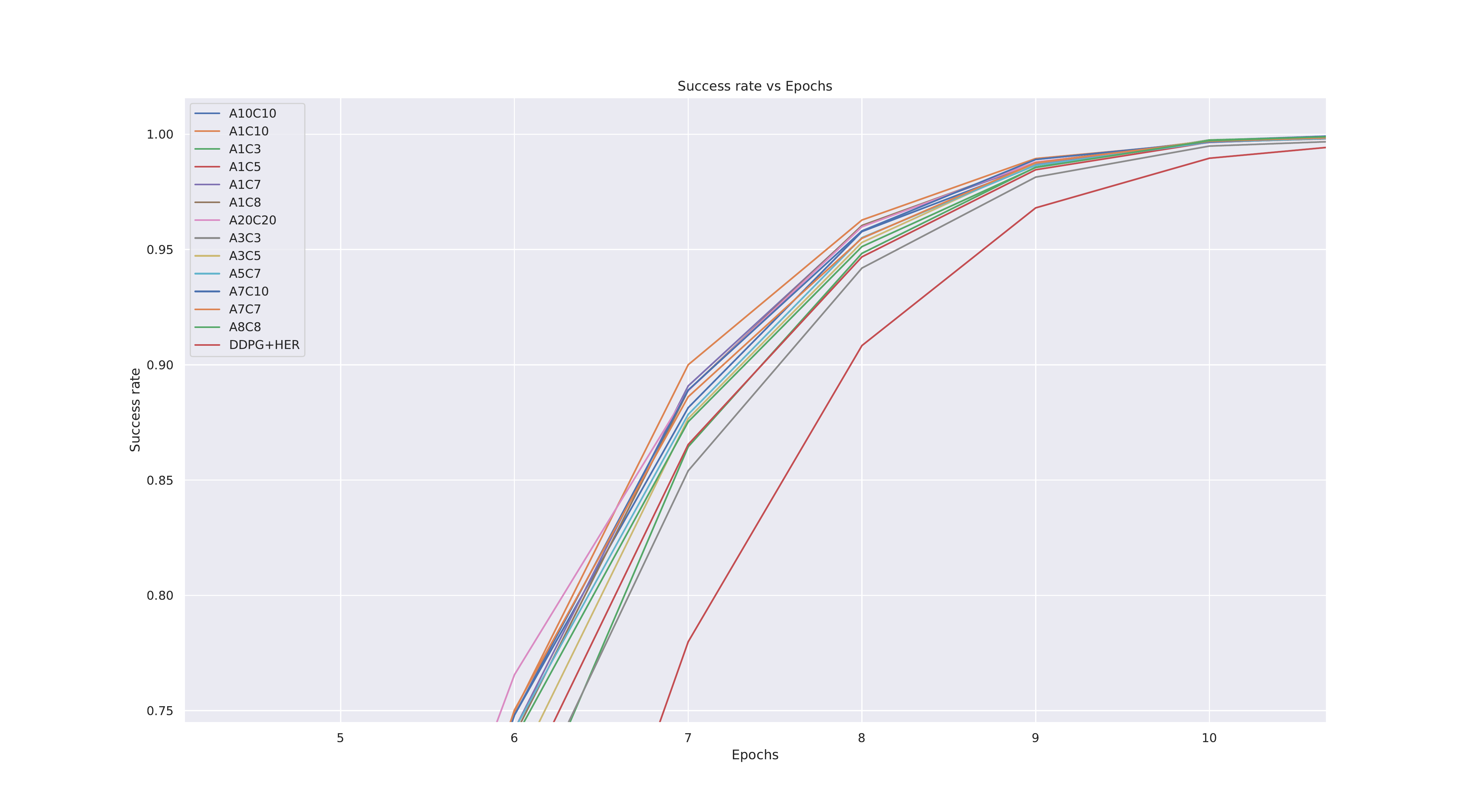}
  \subcaption{Zoomed - Success rate vs Epochs}
  \includegraphics[width=\linewidth,height=4cm]{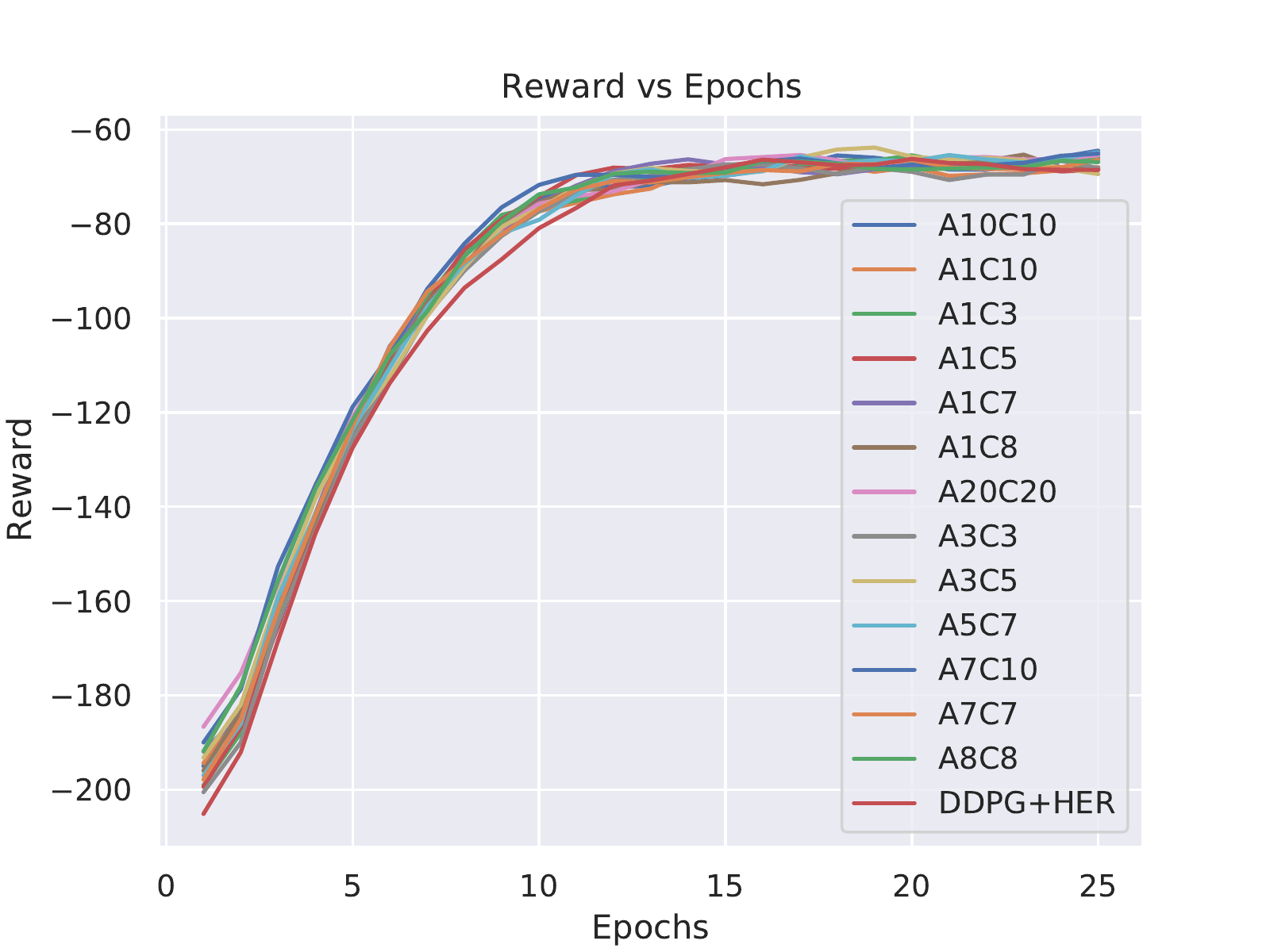}
  \subcaption{Reward vs Epochs}
  \includegraphics[width=\linewidth,height=4cm]{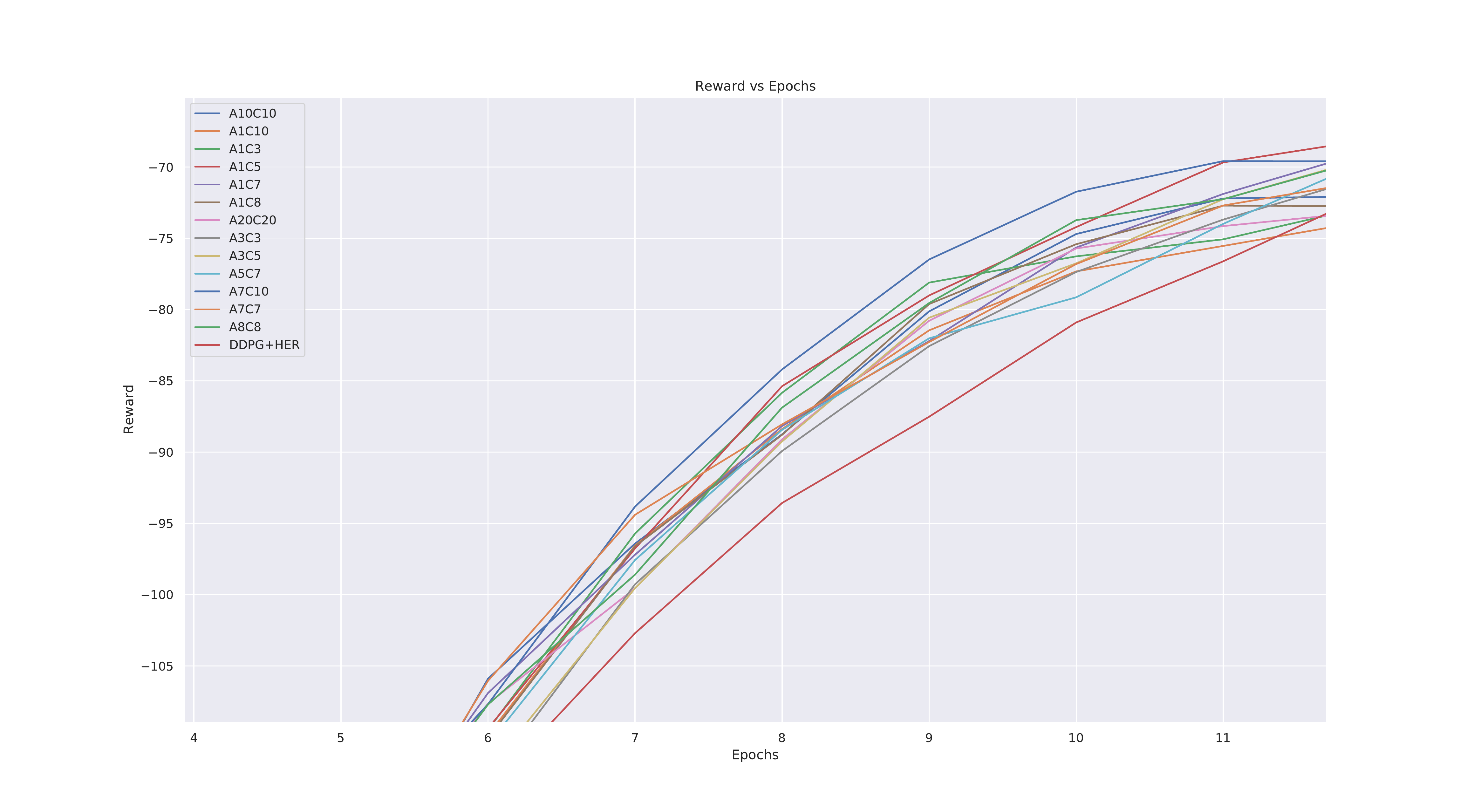}
  \subcaption{Zoomed - Reward vs Epochs}
  \includegraphics[width=\linewidth,height=4cm]{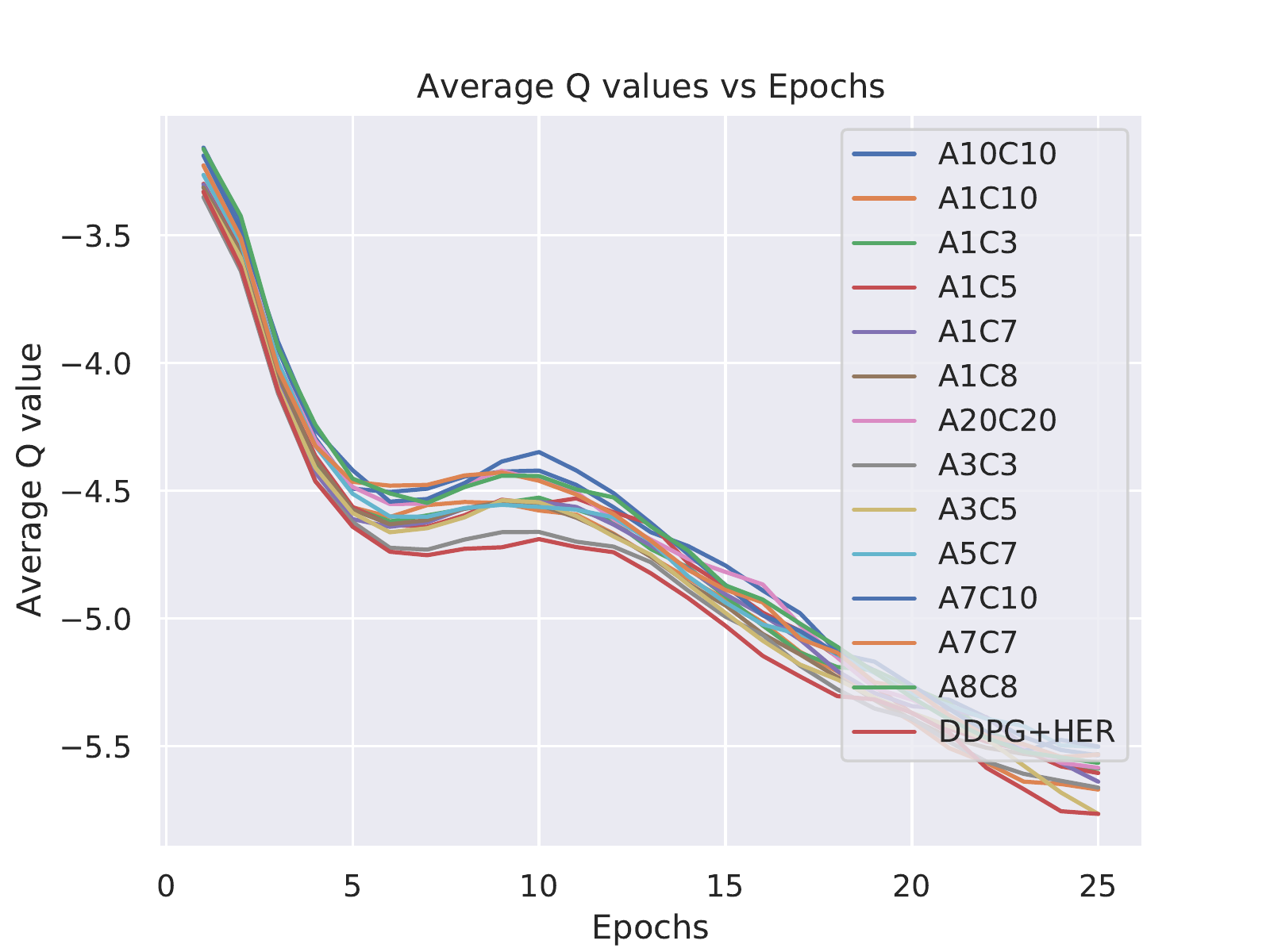}
  \subcaption{Average Q value vs Epochs}
  \includegraphics[width=\linewidth,height=4cm]{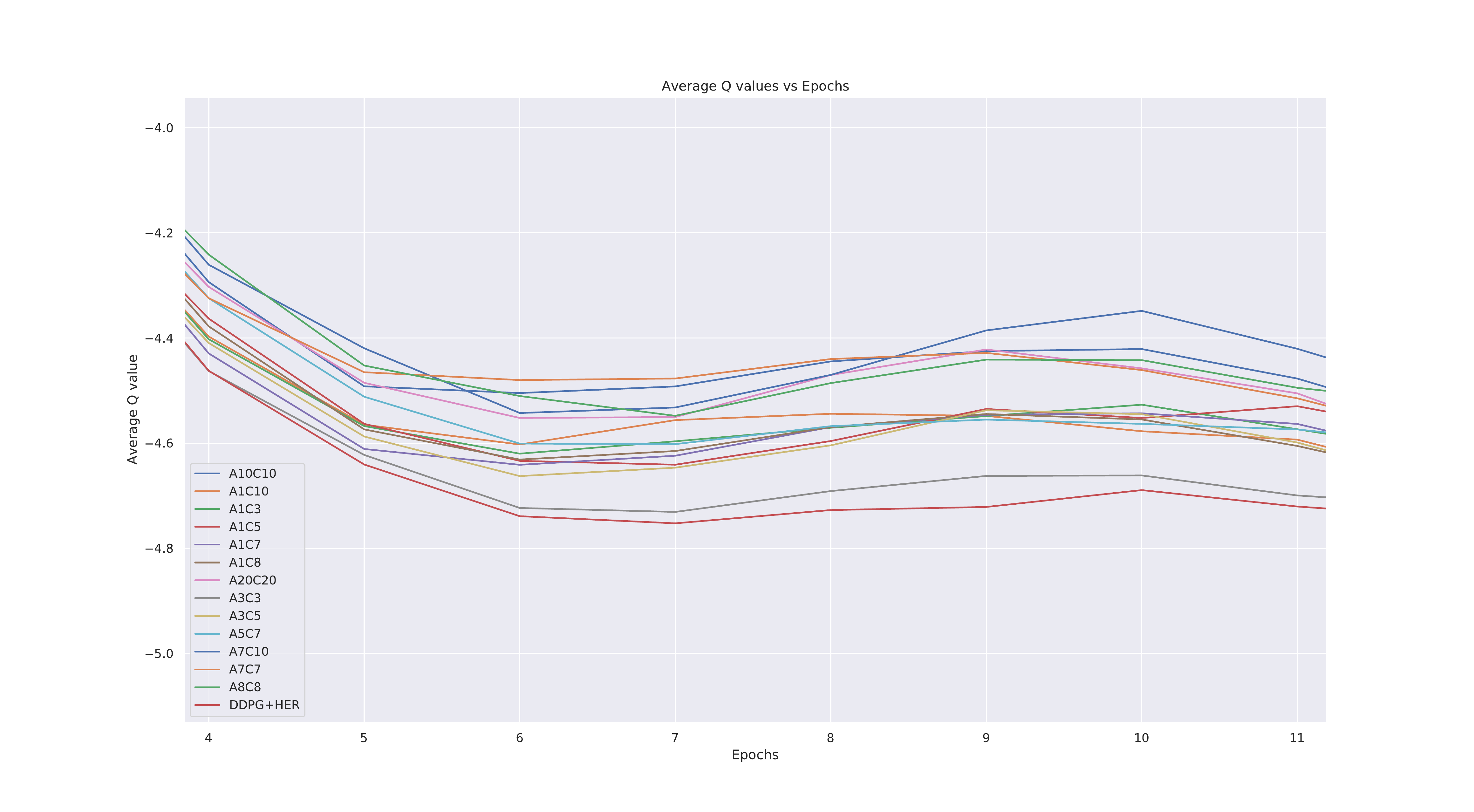}
  \subcaption{Zoomed - Average Q value vs Epochs}
  \end{multicols}
  \caption{Plots comparing different tests carried out utilizing AACHER and DDPG+HER when applied to the \textit{AuboReach} environment. According to the naming convention, each of the AACHER experiments is marked. The success rate, reward, and average Q value have all been compared. These are all plotted against epochs and averaged over 20 runs. For clarity, each plot also displays a corresponding zoomed-in perspective.}
  \label{fig:Plots}
\end{figure*}


\subsection{Simulated Environment}

We are using four Open AI gym environments: \textit{FetchReach-v1}, \textit{FetchPush-v1}, \textit{FetchSlide-v1}, and \textit{FetchPickAndPlace-v1} \cite{brockman2016openai}. In addition to these four environments, we are using the \textit{AuboReach} environment, a custom-built gym environment that we developed for our research. In \textit{AuboReach}, the environment comprises an Aubo i5 manipulator that moves from the starting joint state configuration to the target joint state configuration by following the actions (robot joint states). The configurations of the initial and goal states are arbitrary. Four joints are used in this environment for training and testing (instead of six). The joints used are the \textit{shoulder}, \textit{forearm}, \textit{upper arm}, and \textit{wrist1}. This was done to make sure that the learning could be completed promptly. The range of each joint is -1.7 to 1.7 radians. Figures \ref{fig:auboReachEnv} and \ref{fig:auboReachEnvSim} display real and simulated \textit{AuboReach} settings.


\subsection{Experiment Settings}

\begin{figure*}
\centering
  \begin{multicols}{3}
  \includegraphics[width=\linewidth,height=4cm]{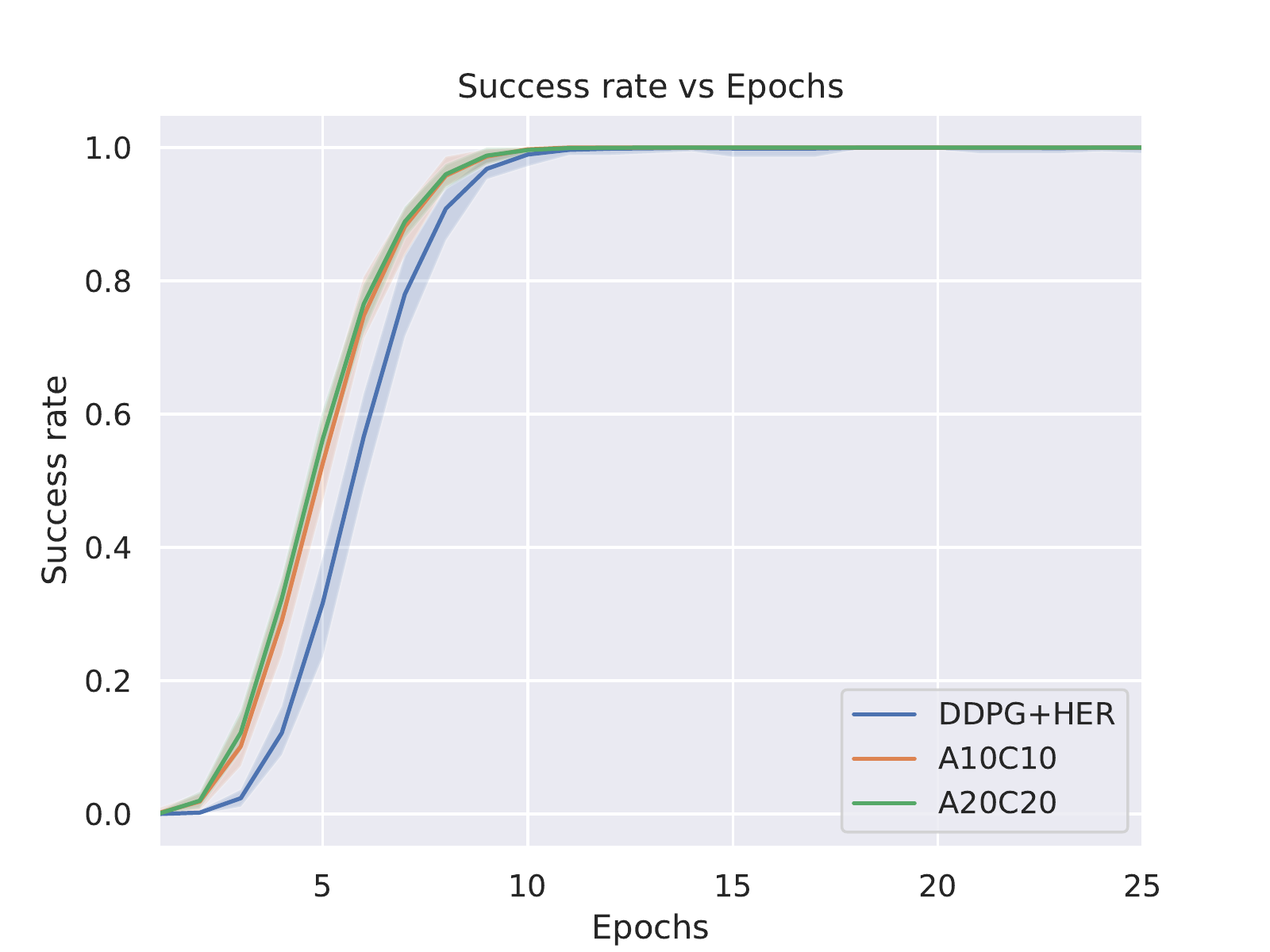}
  \subcaption{Success rate vs Epochs}
  \includegraphics[width=\linewidth,height=4cm]{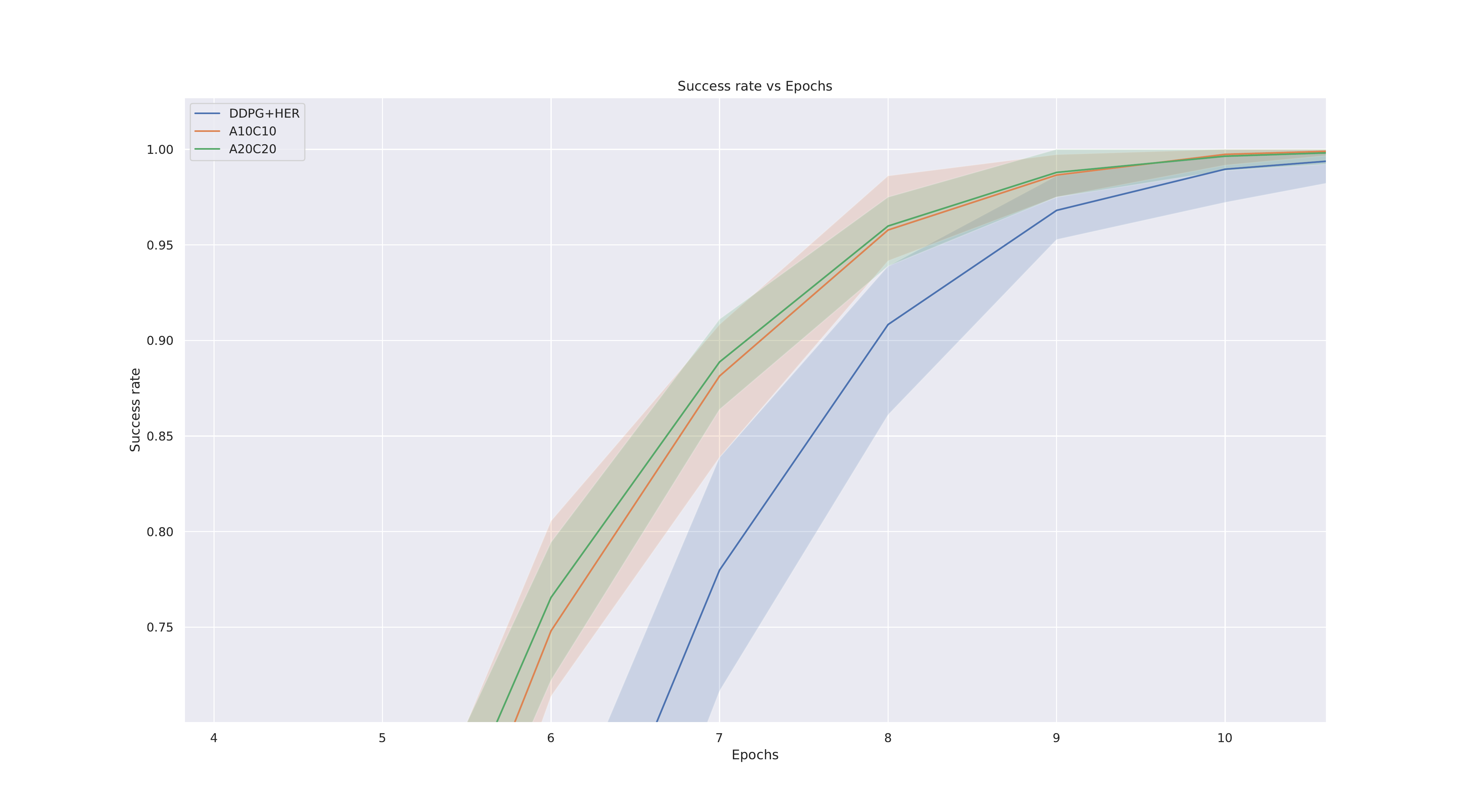}
  \subcaption{Zoomed - Success rate vs Epochs}
  \includegraphics[width=\linewidth,height=4cm]{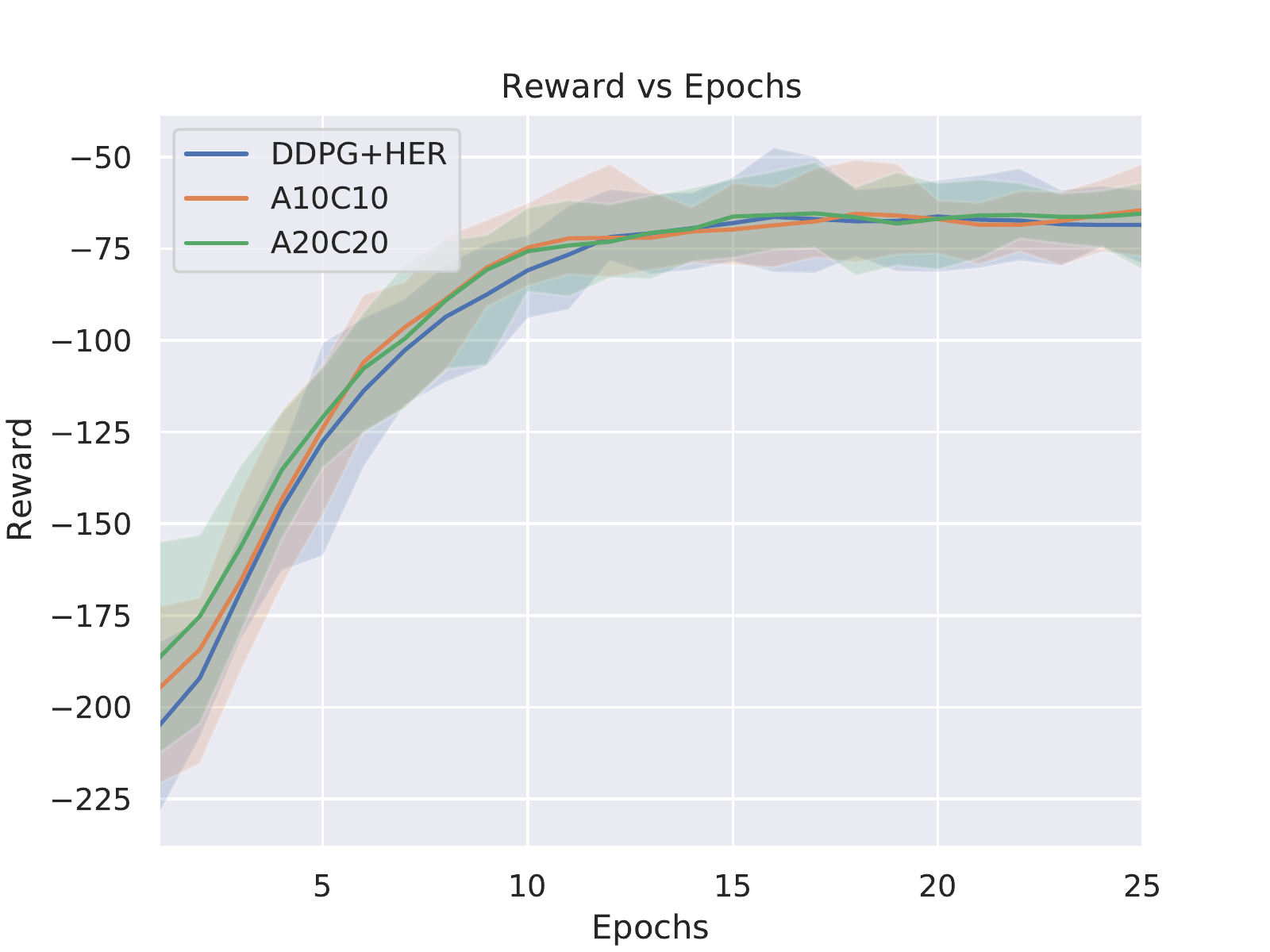}
  \subcaption{Reward vs Epochs}
  \includegraphics[width=\linewidth,height=4cm]{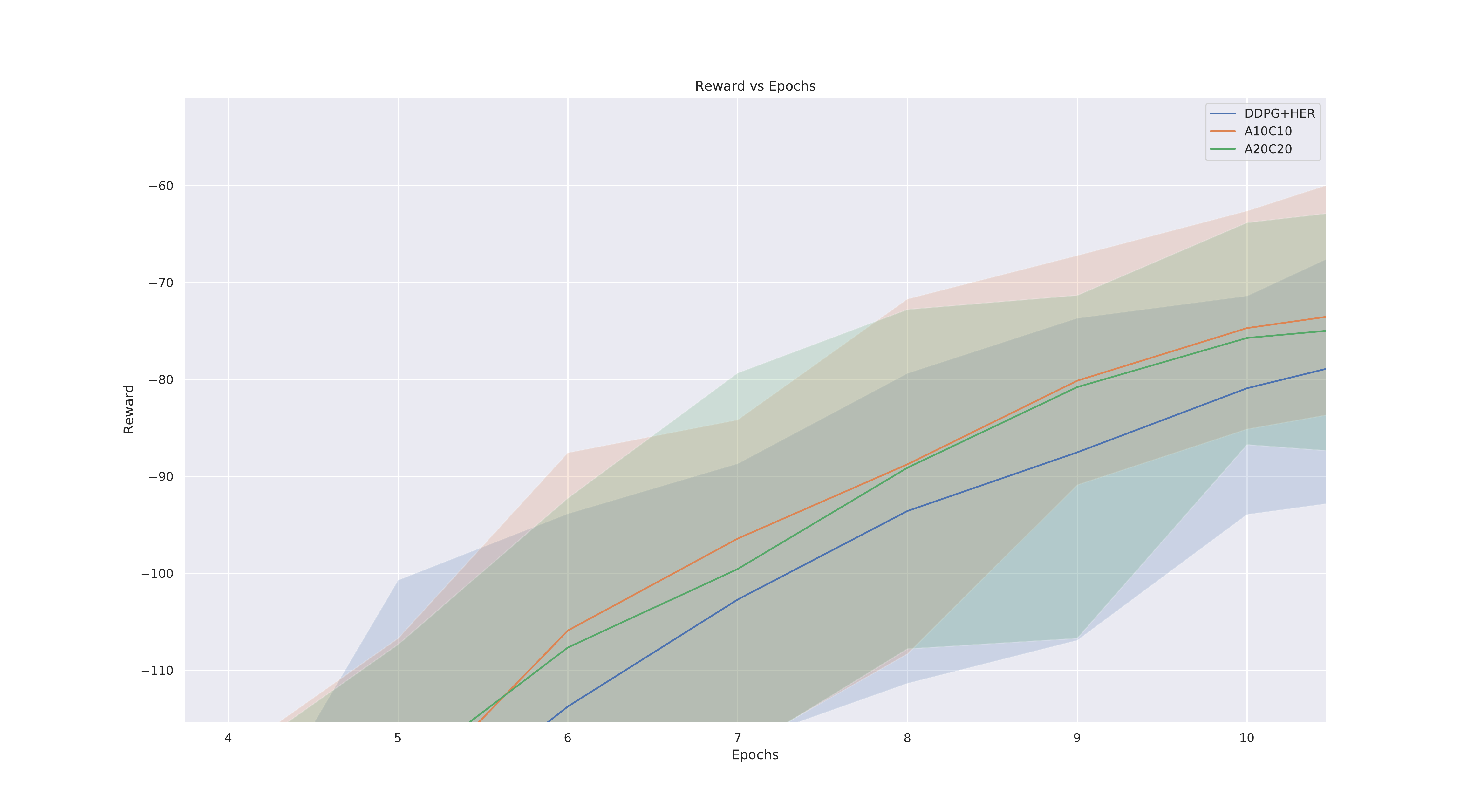}
  \subcaption{Zoomed - Reward vs Epochs}
  \includegraphics[width=\linewidth,height=4cm]{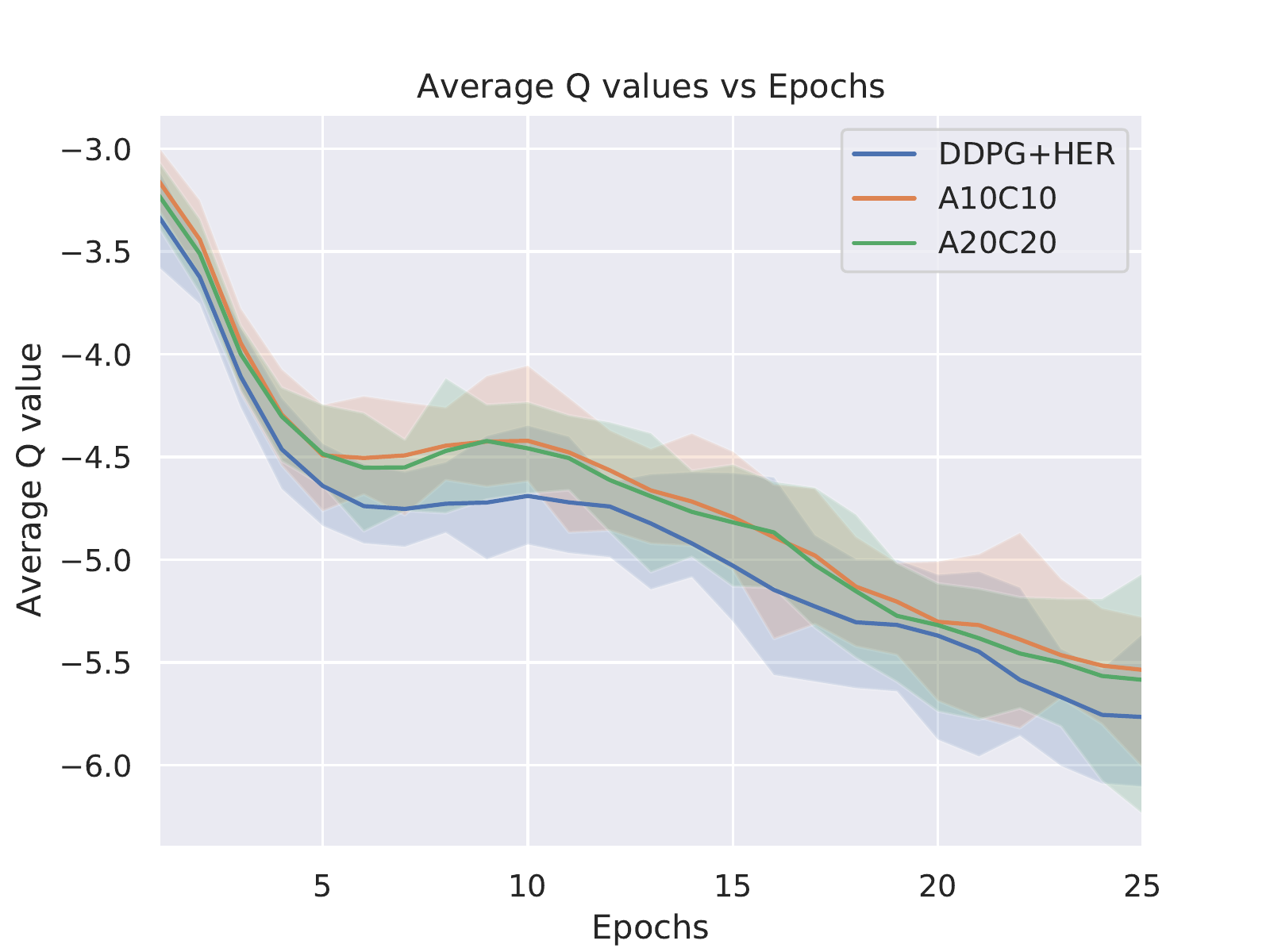}  
  \subcaption{Average Q value vs Epochs}
  \includegraphics[width=\linewidth,height=4cm]{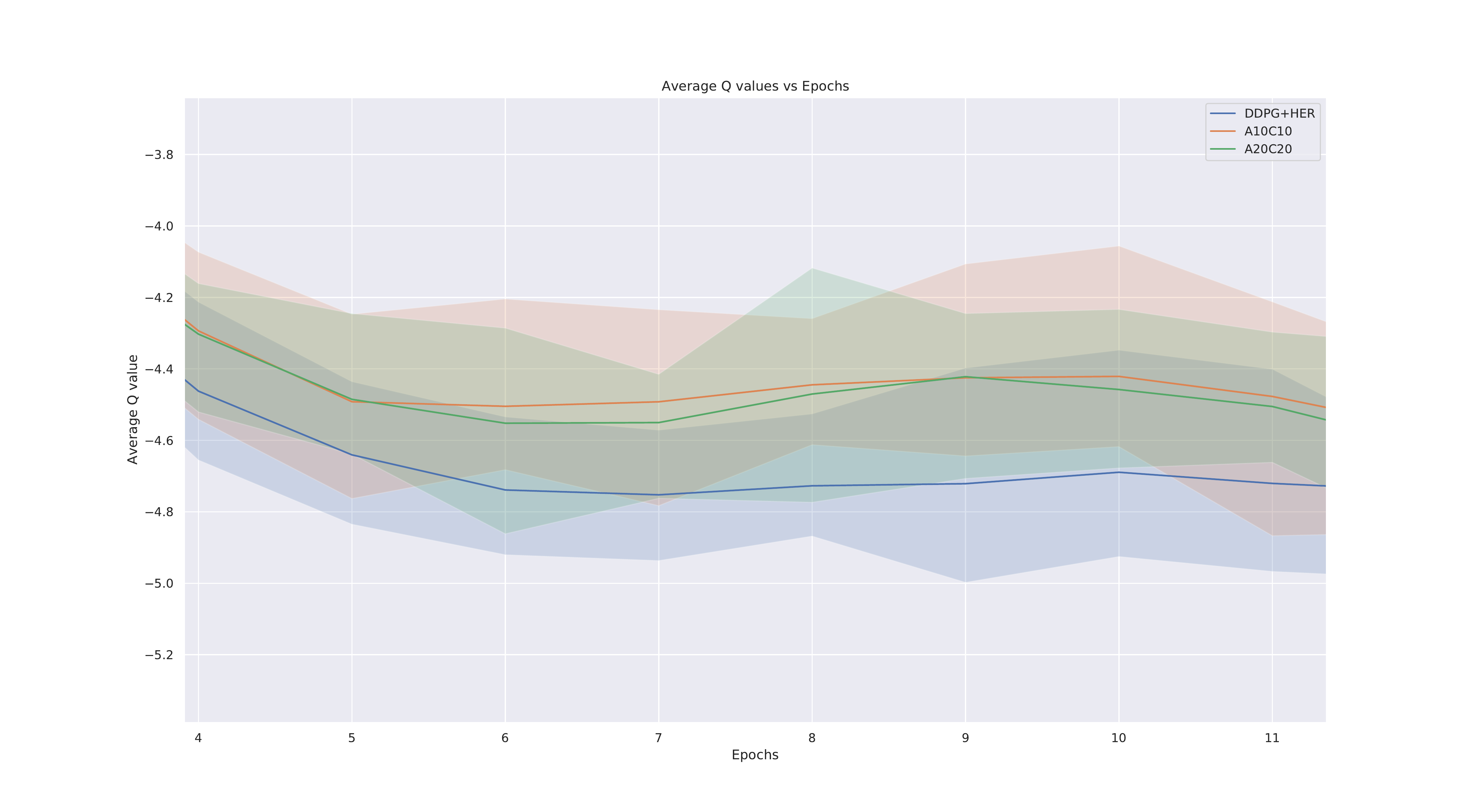}
  \subcaption{Zoomed - Average Q value vs Epochs}
  \end{multicols}
  \caption{The two top-performing experiments, A10C10 and A20C20, when applied to the \textit{AuboReach} environment, are plotted against DDPG+HER for comparison. Using epochs and an average of 20 runs, plots for success rate, reward, and average Q values are displayed. The shaded region displays the range of values for each plot throughout 20 runs. A zoomed-in version of each plot is included for clarity.}
  \label{fig:plots_2_aubo}
\end{figure*}

The specific settings for each experiment are covered in this subsection. Every experiment has certain predetermined parameters. Actor and critic learning rates are set at 0.001. Target networks' update rates are 0.01, and the discount factor is 0.98. A zero-mean Gaussian noise with a variance of 0.2 is added to the action during the exploration. The training method for each experiment consists of 25 epochs, each of which comprises 15 cycles. Every cycle, the robot rolls out 100 steps, followed by 20 times robot training. The default batch size is 256. The experience replay buffer is designed as a $10^6$ length circular queue. All experience tuples (state, action, reward, and next state) are saved during each trajectory roll-out and kept in a replay buffer of finite size. When we update the value and policy networks, we sample random mini-batches of experience from the replay buffer. The number of additional goals used for the replay is $k$=4, which means that the replay buffer will immediately retain 20\% of normal transitions with the original goal. When neural network weights grow incredibly big, L2 regularization is added to the loss. Both actor and critic networks employ layer normalization \cite{ba2016layer}. The robot's observations are also normalized. Adam \cite{kingma2014adam}, a momentum-based method, is utilized to optimize the loss function during training. We are using Ubuntu 16.04 and a GeForce GTX 1080 Ti Graphic card for our experiment.

\begin{figure*}[!h]
\centering
\begin{multicols}{4}
    \includegraphics[width=\linewidth,height=3.5cm]{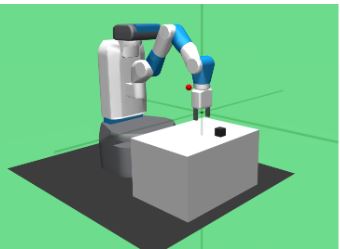}
    \subcaption{FetchPick\&Place environment}
    \includegraphics[width=\linewidth,height=4cm]{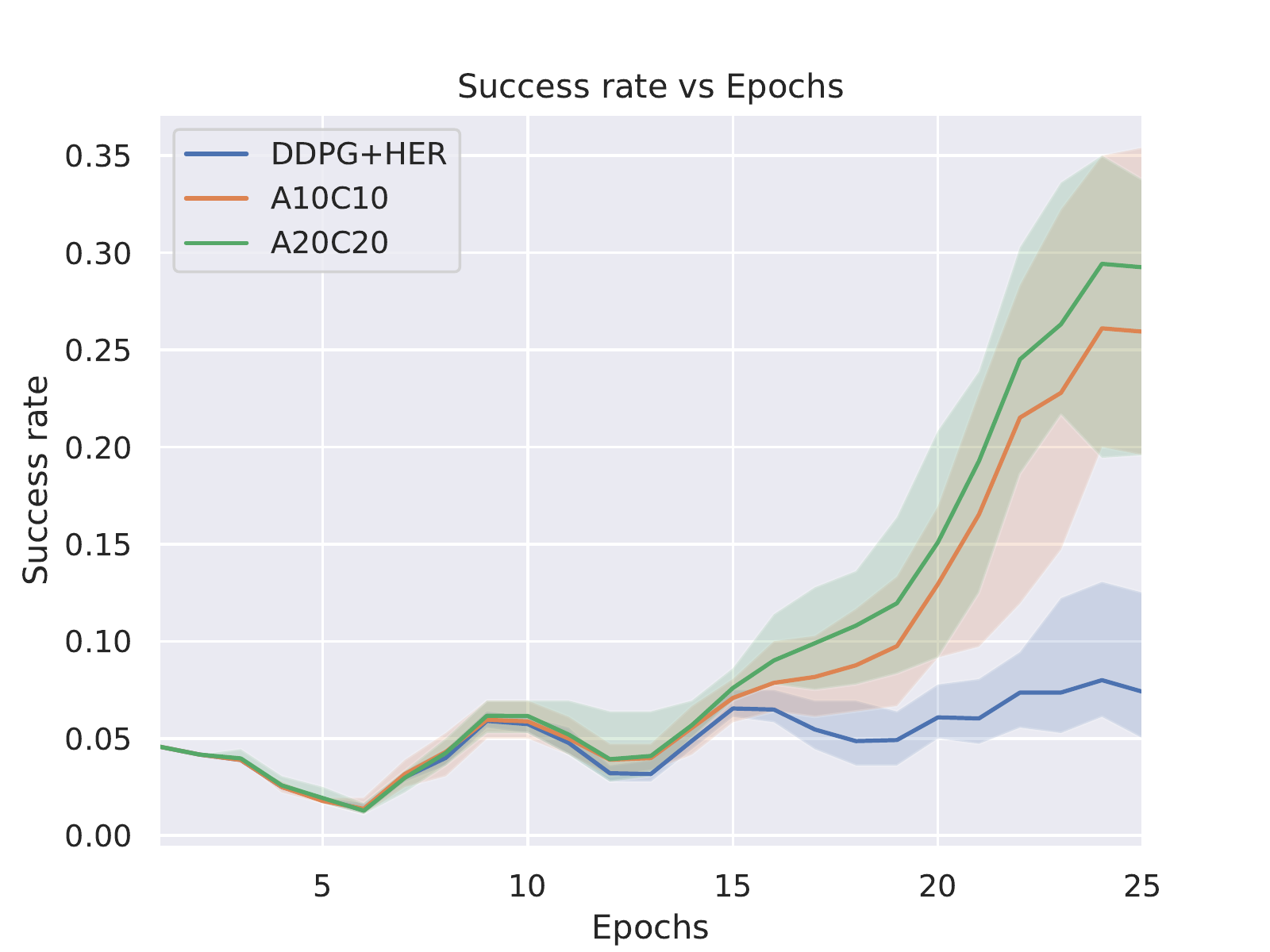}
    \subcaption{Success rate vs Epochs}
    \includegraphics[width=\linewidth,height=4cm]{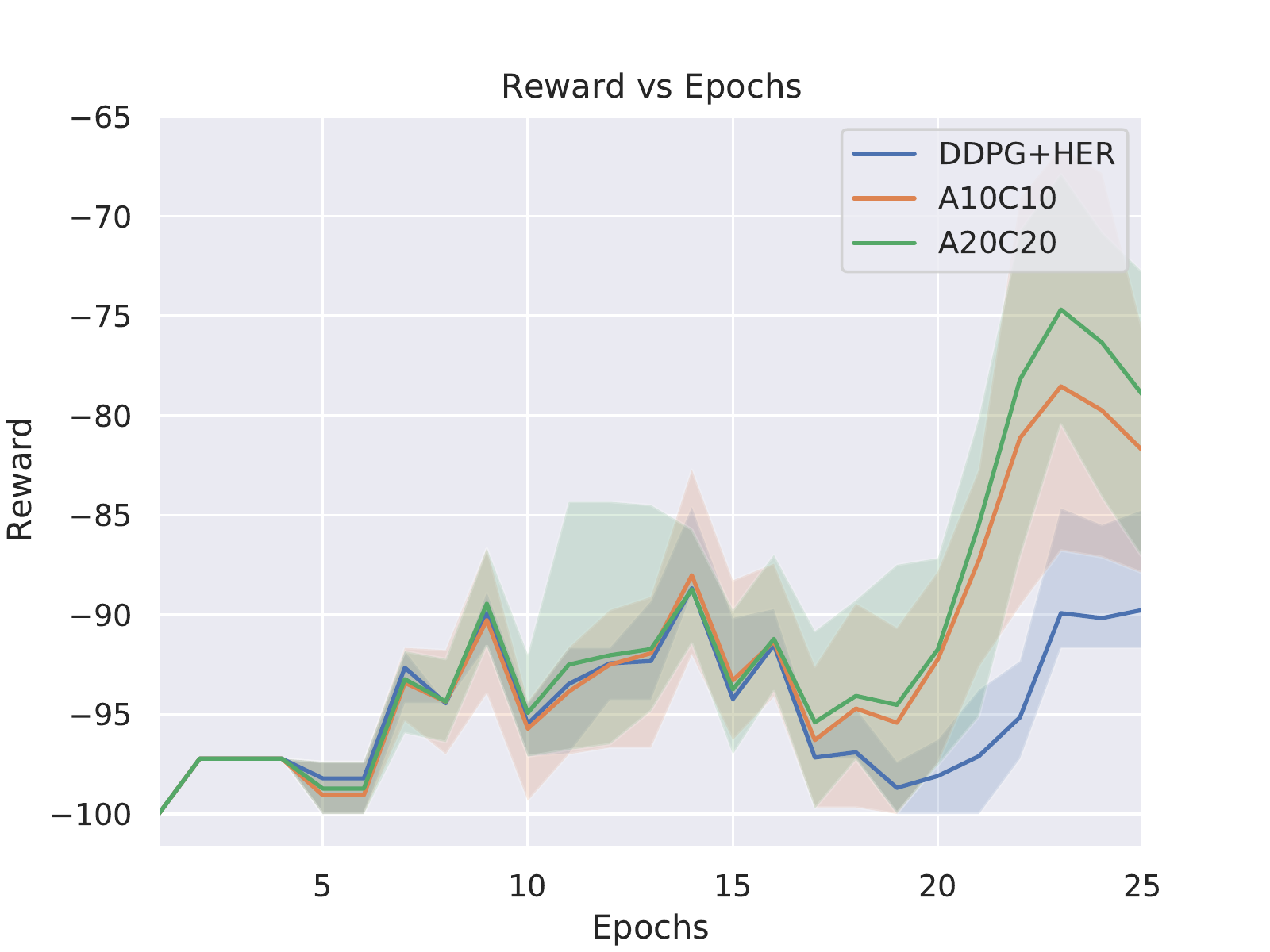}
    \subcaption{Reward vs Epochs}
    \includegraphics[width=\linewidth,height=4cm]{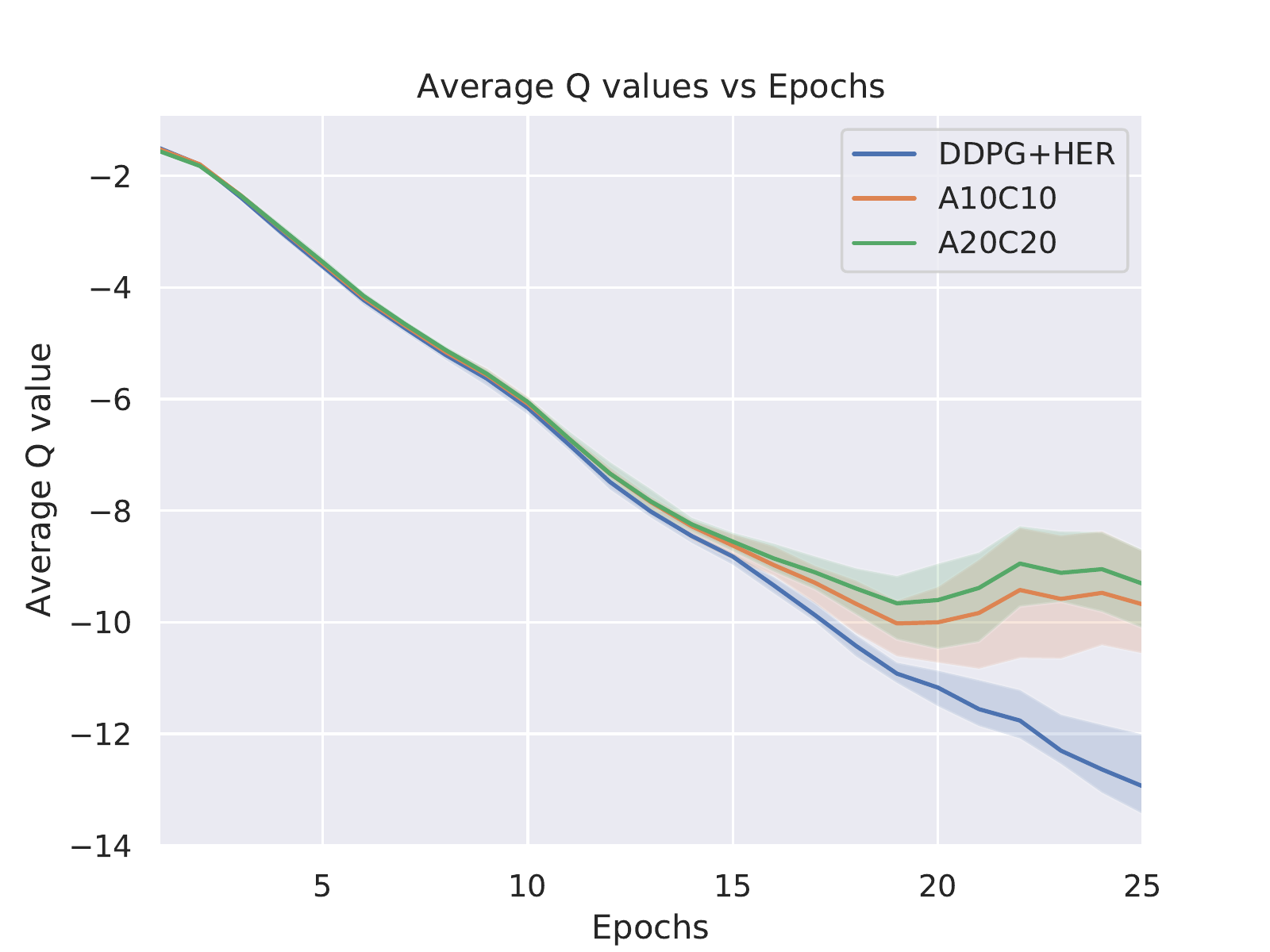}
    \subcaption{Average Q value vs Epochs}
    \includegraphics[width=\linewidth,height=3.5cm]{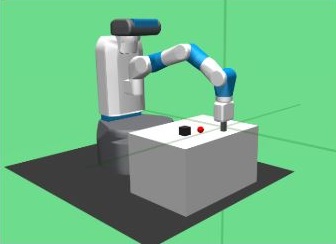}
    \subcaption{FetchPush environment}
    \includegraphics[width=\linewidth,height=4cm]{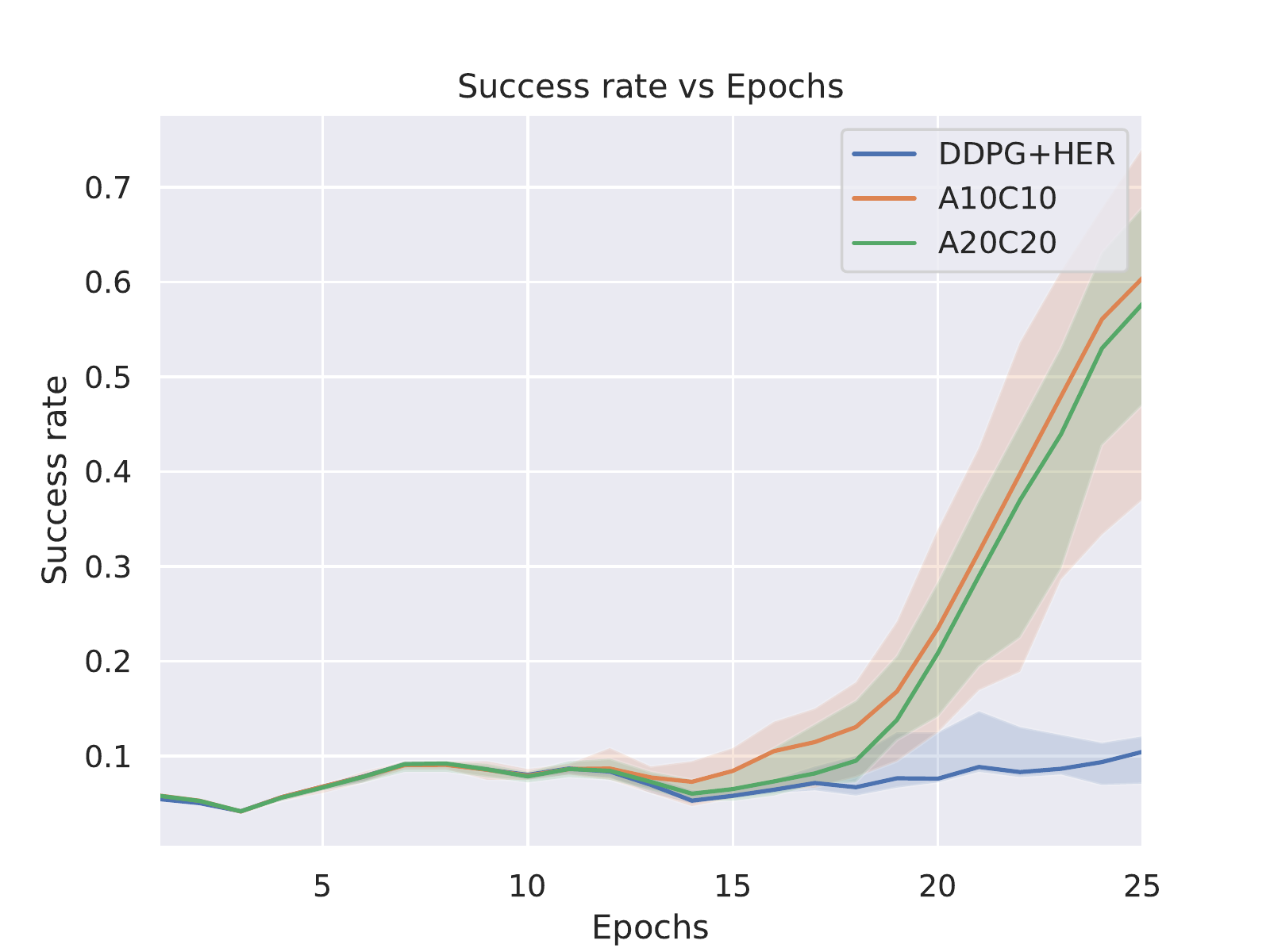}
    \subcaption{Success rate vs Epochs}
    \includegraphics[width=\linewidth,height=4cm]{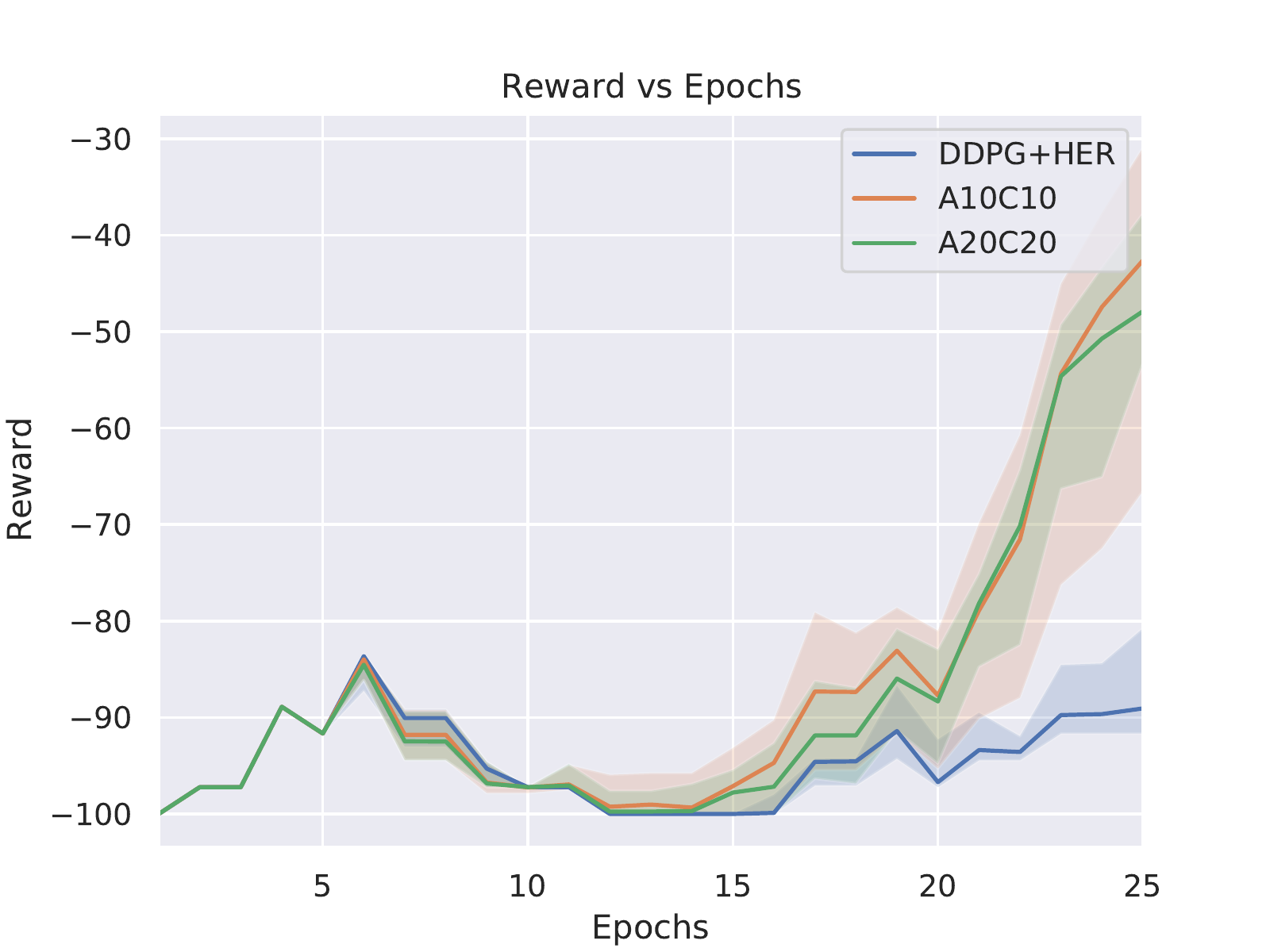}
    \subcaption{Reward vs Epochs}
    \includegraphics[width=\linewidth,height=4cm]{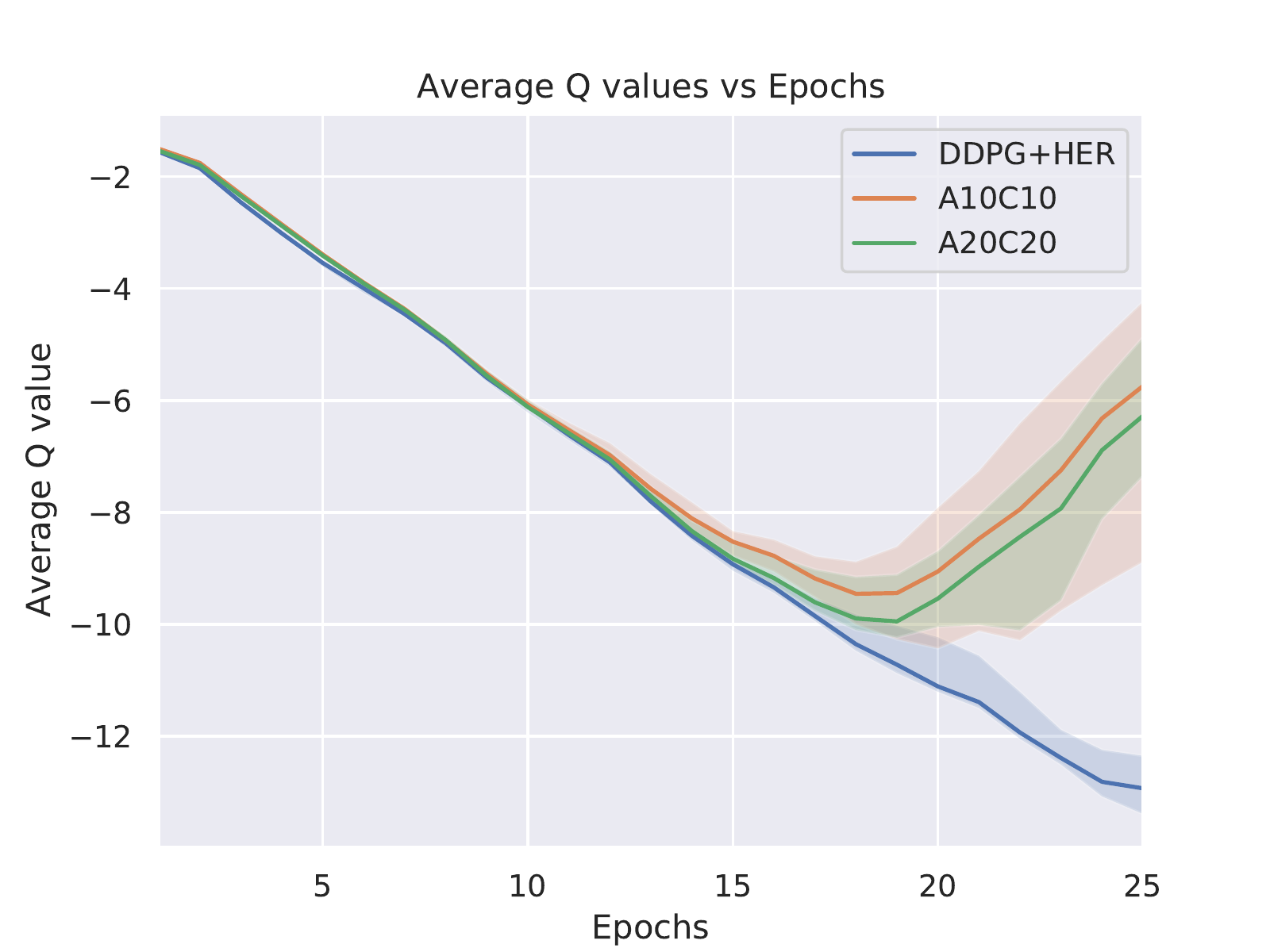}
    \subcaption{Average Q value vs Epochs}
    \includegraphics[width=\linewidth,height=3.5cm]{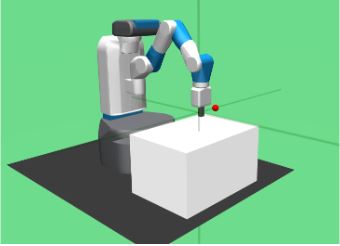}
    \subcaption{FetchReach environment}
    \includegraphics[width=\linewidth,height=4cm]{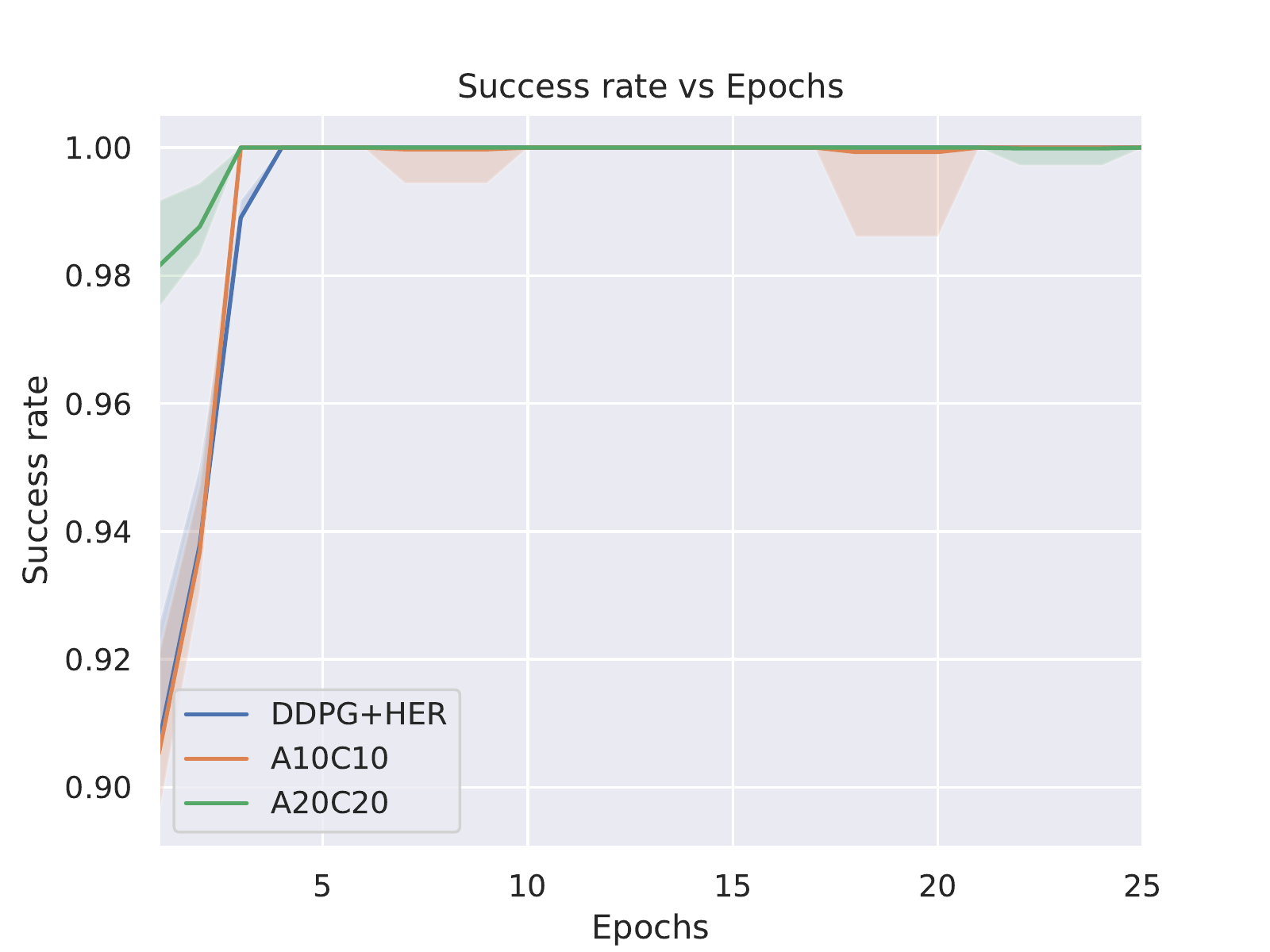}
    \subcaption{Success rate vs Epochs}
    \includegraphics[width=\linewidth,height=4cm]{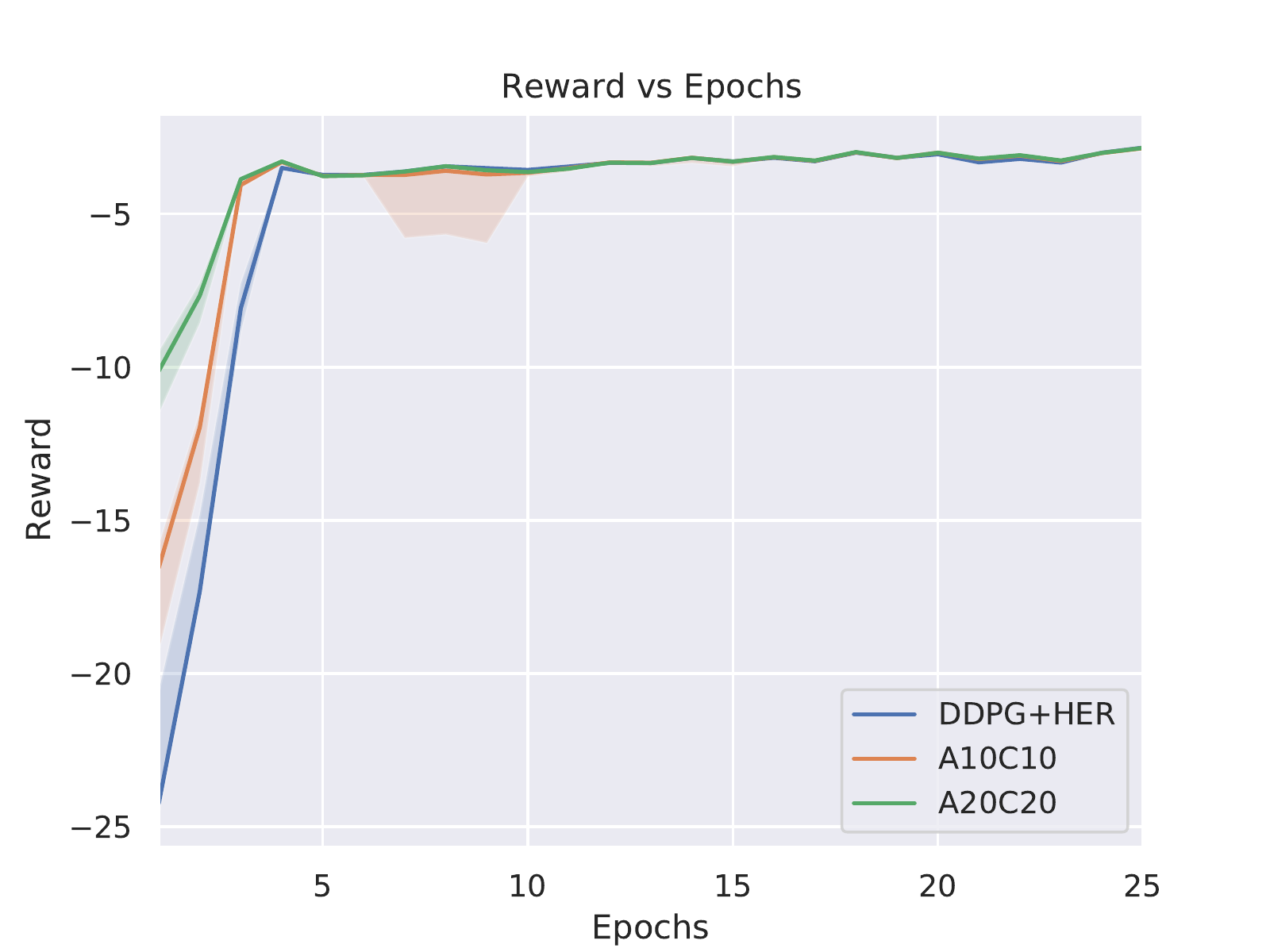}
    \subcaption{Reward vs Epochs}
    \includegraphics[width=\linewidth,height=4cm]{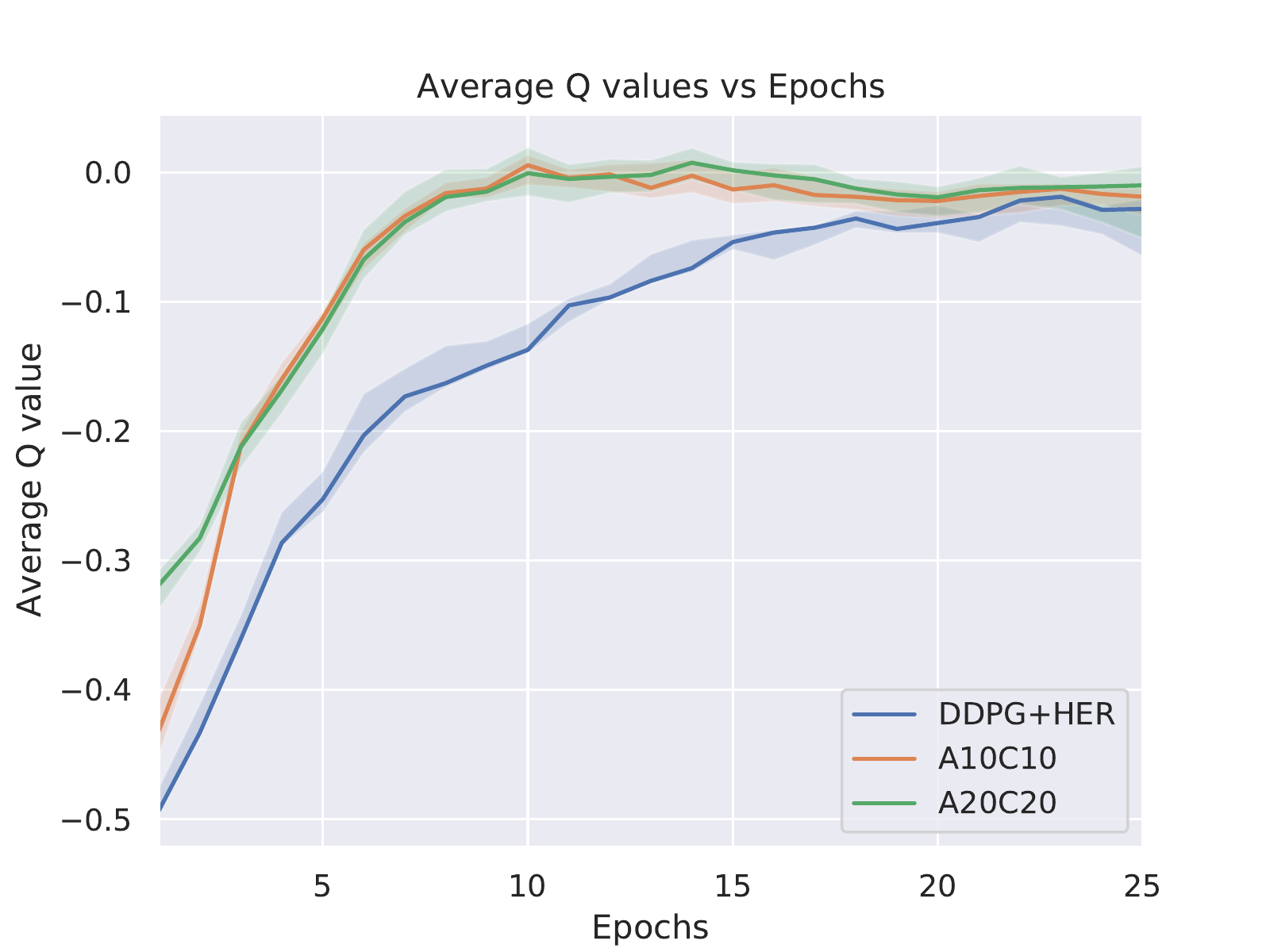}
    \subcaption{Average Q value vs Epochs}
    \includegraphics[width=\linewidth,height=3.5cm]{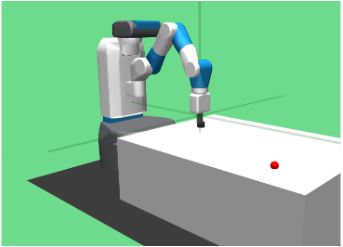}
    \subcaption{FetchSlide environment}
    \includegraphics[width=\linewidth,height=4cm]{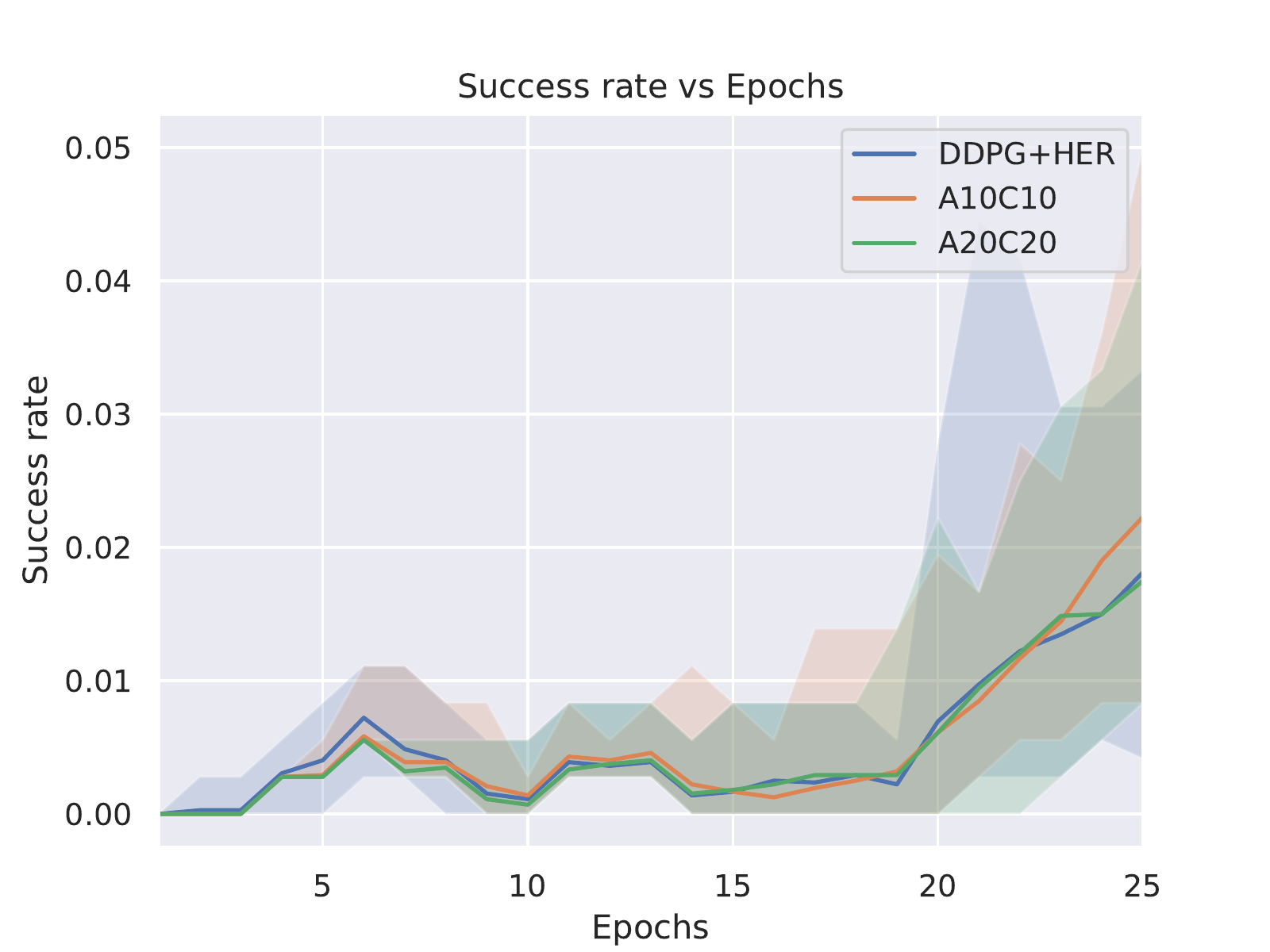}
    \subcaption{Success rate vs Epochs}
    \includegraphics[width=\linewidth,height=4cm]{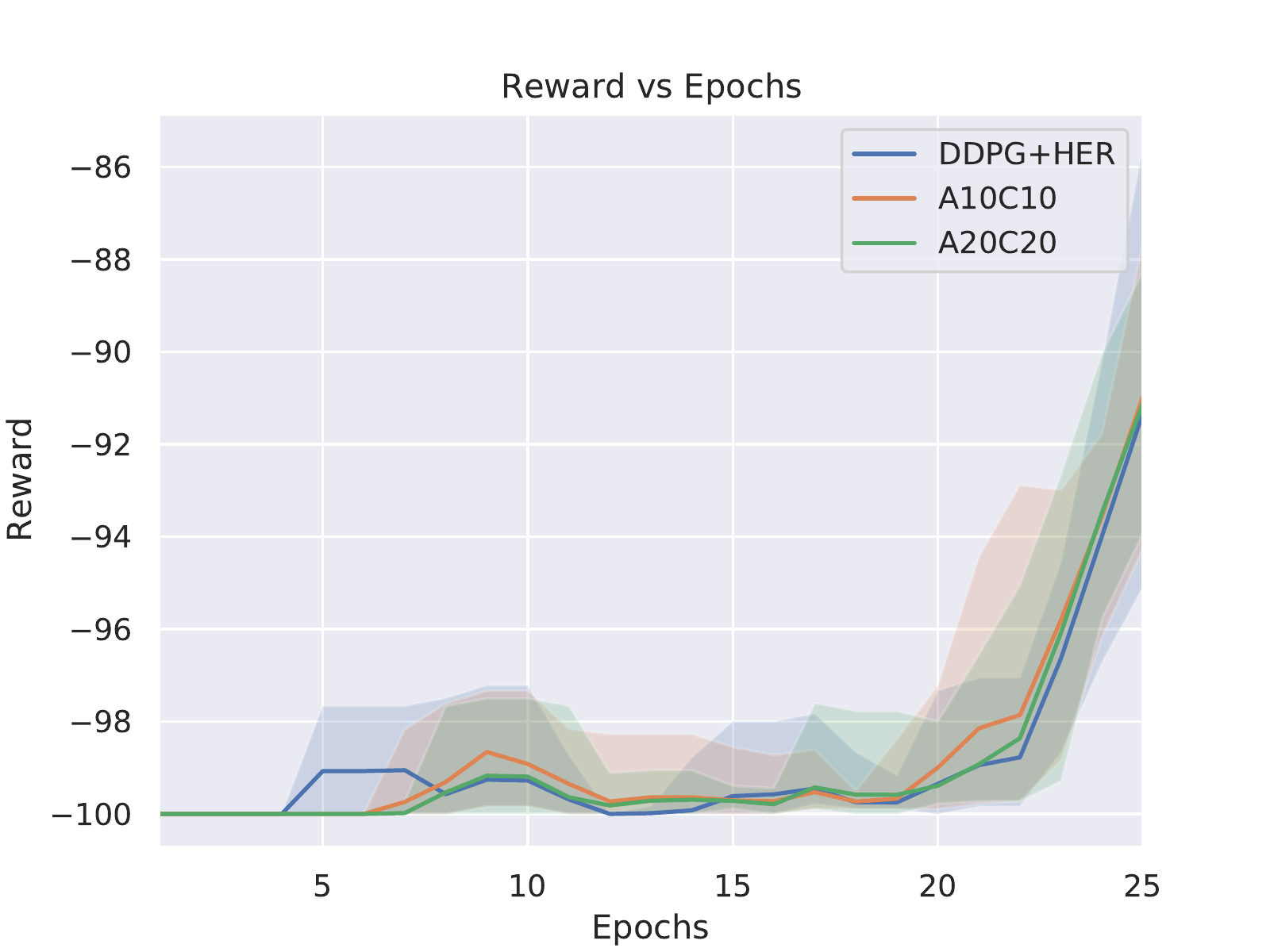}
    \subcaption{Reward vs Epochs}
    \includegraphics[width=\linewidth,height=4cm]{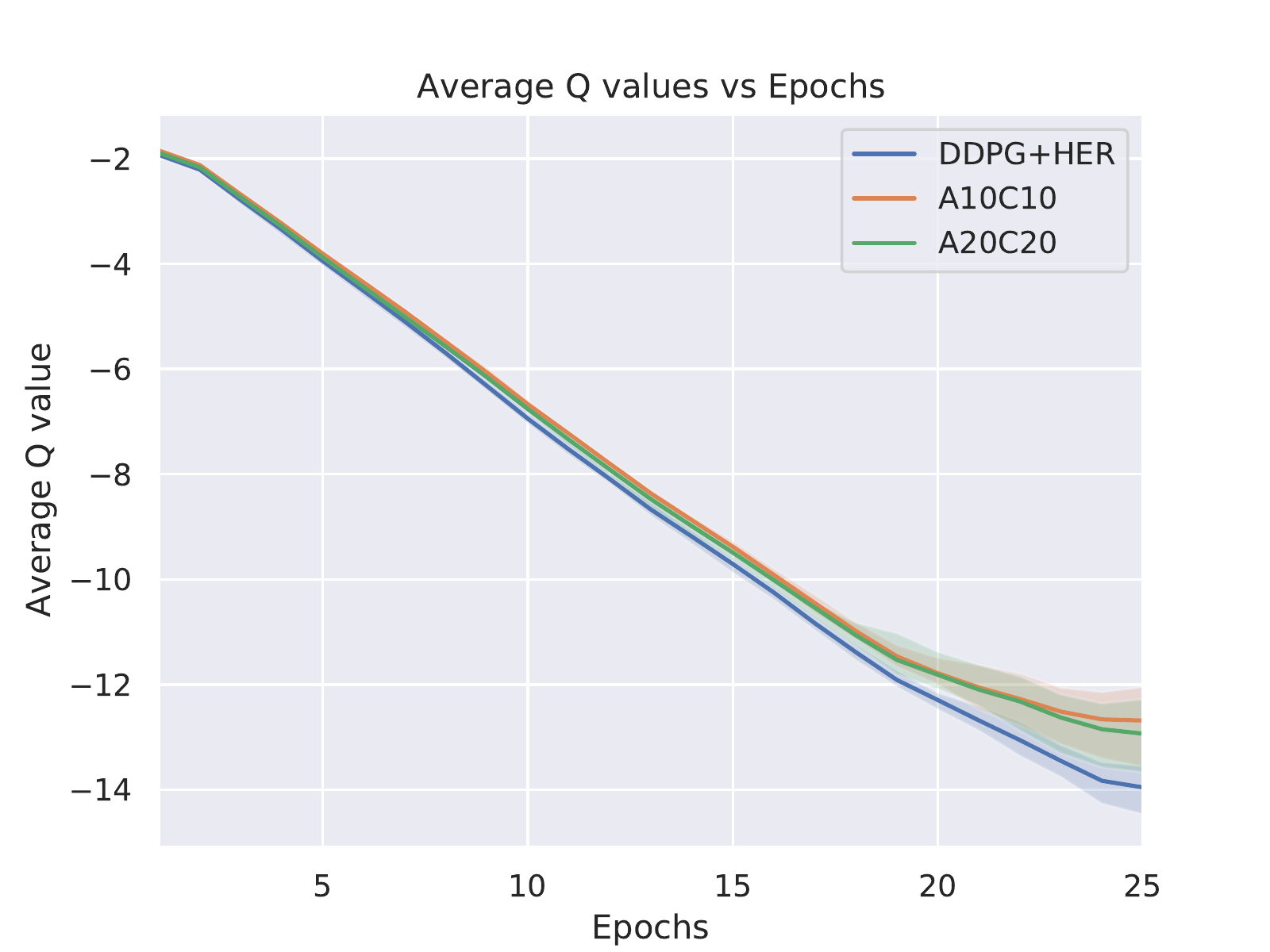}
    \subcaption{Average Q value vs Epochs}
\end{multicols}
\caption{Plotted against DDPG+HER for comparison are the two best-performing trials, A10C10 and A20C20 when applied to the various settings. Plots for each habitat are displayed beneath them. Plots for success rate, reward, and average Q values are shown using epochs and an average of 20 runs. The range of values for each plot throughout the 20 runs is shown in the shaded area.}
\label{finalPlots}
\end{figure*}

In every experiment, the actor and the critic have three hidden layers, each with 256 units. AACHER tests out several combinations of actor and critic networks. The actor and critic networks both have similar shapes. To ensure accurate experiment findings, DDPG+HER and AACHER were each separately tested 20 times.

To prevent the robot from running into objects in its environment and to make the training process easier, training was done in a simulated \textit{AuboReach} environment. Additionally, the robot's motors were disabled during simulation training. This means that it is assumed that the robot can successfully transition to whatever action (set of joint states) is selected by any of the algorithms being tested.  This environment's implementation involves the use of \textit{MOVEit} package \cite{hernandez2017design}, which takes care of the planning and execution required for internal joint movement in the robot \cite{field1996iterative}. Additionally, by following these steps, the chance of a collision is avoided, which might have happened otherwise. Hence, it is safe to assume that the robot will have no collision within its own joints.

\textit{AuboReach} rates the algorithm-determined joint states as successful if the total deviation between the target and the attained joint states is less than 0.1 radians. The objective joint states for training were [-0.503, 0.605, -1.676, 1.391]. The random initial and target state configurations were tested in actual and virtual situations with motors turned on using the policy generated by training in AuboReach. The robot was able to transition between user-selected initial and target state configurations successfully.

\subsection{Experiment Results and Discussions}

The tests for the DDPG+HER and AACHER methods are all presented in this subsection. They were all carried out in the previously specified simulated environment. Additionally, impartial assessments of these outcomes are performed.

By simulating the \textit{AuboReach}, \textit{FetchReach-v1}, \textit{FetchPush-v1}, \textit{Fetch Slide-v1}, and \textit{FetchPickAndPlace-v1} environments, we conducted experiments using the methodologies. Specifically, for \textit{AuboReach}, to avoid any unexpected manipulator movements, a virtual environment was used for training. Following the training, the policy file was tested using a real \textit{AuboReach} manipulator setup. Even though no measurements were produced for the real setup, it confirms the trained policy's efficacy.



The DDPG+HER and AACHER methods are assessed in the \textit{AuboReach}, \textit{FetchReach-v1}, \textit{FetchPush-v1}, \textit{FetchSlide-v1}, and \textit{FetchPick AndPlace-v1} environments. For evaluation, we are using these three matrices:  average Q values, success rate, and reward. Each plot is plotted against epochs and averaged across 20 runs. We first evaluated our algorithm in the \textit{AuboReach} environment. Plots in Figure \ref{fig:Plots} demonstrate our results in the \textit{AuboReach} environment and show that AACHER outperforms DDPG+HER in terms of performance. This is demonstrably supported by the finding that the success rate of AACHER trials is higher than that of DDPG+HER studies. The reward and average Q values are in a comparable situation. This validates the stability and efficacy of the AACHER technique. The two best-performing A10C10 and A20C20 were selected for further comparison and plotted against DDPG+HER in a separate Figure \ref{fig:plots_2_aubo}. The shaded regions with color denote the ranges of maximum and minimum values for all experiment findings in the figures. In all three matrices, it is observed that A10C10 and A20C20's performance is pretty similar to one another. Our algorithm exhibits better performance than DDPG+HER. 

Finally, we evaluated our algorithm using the remaining four simulated environments, \textit{FetchReach-v1}, \textit{FetchPush-v1}, \textit{FetchSlide-v1}, and \textit{FetchPickAndPlace-v1}. We only used the top-performing A10C10 and A20C20 from the \textit{AuboReach} environment in comparison to DDPG+HER for further evaluations in these environments. Figure \ref{finalPlots} shows all the plots for these environments for average Q values, success rate, and reward.

Figure \ref{finalPlots} shows the overall performance of our proposed algorithm, AACHER, which outperforms the conventional DDPG+HER when applied to OpenAI's gym environments. In the case of the \textit{FetchPickAndPlace-v1} environment, the performance of A20C20 is found to be the best in all three metrics. While DDPG+HER's performance could not significantly improve, that of A10C10 is comparable to that of A20C20. The performance of A10C10 is found to be superior to others in all matrices in the \textit{FetchPush-v1} environment. A20C20 performs nearly as well as A10C10; however, DDPG+HER fell short. The three approaches in \textit{FetchReach-v1} have similar success rates and rewards, but A10C10 and A20C20 have much higher average Q values. A10C10 surpasses the others across all matrices, even though all of the methods perform fairly similarly in the \textit{FetchSlide-v1} environment, which is thought to be a difficult task.

\begin{table}[ht]
\centering
\begin{tabular}{|p{0.1\linewidth}|p{0.11\linewidth}|p{0.1\linewidth}|p{0.1\linewidth}|p{0.1\linewidth}|p{0.1\linewidth}|p{0.1\linewidth}|} 
 \hline
 {} & \textbf{Setting} & \textbf{Aubo Reach} & \textbf{Fetch Reach-v1} & \textbf{Fetch Push-v1} & \textbf{Fetch Slide-v1} & \textbf{Fetch Pick And Place-v1} \\
 \hline
 \textbf{Success rate} & \textbf{DDPG+ HER} & 1 & 1 & 0.13 & 0.016 & 0.08\\
 \textbf{} & \textbf{A10C10} & 1 & 1 & \textbf{0.652} & \textbf{0.023} & 0.309\\ 
 \textbf{} & \textbf{A20C20} & 1     & 1 & 0.647 & 0.013 & \textbf{0.339}\\ 
 \hline
 \textbf{Rewar -d} & \textbf{DDPG+ HER} & -69.3 & -2.5 & -89.6 & -90.81 & -96.9\\
 \textbf{} & \textbf{A10C10} & \textbf{-63.00} & -2.5 & \textbf{-45.00} & \textbf{-90.70} & -83.4\\ 
 \textbf{} & \textbf{A20C20} & -64.60 & -2.5 & -51.83 & -90.8 & \textbf{-81.04}\\ 
 \hline
 \textbf{Avera -ge Q value} & \textbf{DDPG+ HER} & -5.8 & -0.04 & -12.84 & -14.16 & -12.9\\
 \textbf{} & \textbf{A10C10} & \textbf{-5.49} & -0.02 & \textbf{-5.016} & \textbf{-12.74} & -9.19\\ 
 \textbf{} & \textbf{A20C20} & -5.66 & \textbf{-0.01} & -5.19 & -13.06 & \textbf{-8.808}\\ 
 \hline
\end{tabular}
\caption{For each of the five environments, the success rate, reward, and average Q value are shown in the table. The average of all the values across 20 runs is for the $25^{th}$ epoch.}
\label{table:drl1}
\end{table}

Table \ref{table:drl1} provides a summary of the experiment's findings. The average of every value, including the success rate, reward, and average Q values, is shown. Each number in the table is an average of over 20 runs during the $25^{th}$ epoch. The numbers in bold indicate the best outcomes for a certain statistic in a particular setting. Robots in some of the environments show a success rate of 1, as seen in the table. The robot looks to be still learning in \textit{FetchPush-v1}, \textit{FetchSlide-v1}, and \textit{FetchPickAndPlace-v1}, and the success rate is still well below the maximum success rate of 1. A10C10 has the best success rate in \textit{FetchPush-v1} and \textit{FetchSlide-v1}, whereas \textit{FetchPickAndPlace-v1} has the highest success rate for A20C20. The rise in the success rate of A10C10 at the $25^{th}$ epoch is almost 3.8 times more than that of DDPG+HER in \textit{FetchPush-v1}. In \textit{FetchPickAndPlace-v1}, A20C20's improvement in success rate is around 3.3 times larger than DDPG+HER's. In the \textit{FetchPush-v1} environment, the rewards for A10C10 are 50\% higher than those for DDPG+HER. The Average Q value for \textit{FetchReach-v1} for A20C20 is 76\% greater than DDPG+HER. This indicates the superior efficiency of our algorithm, AACHER, over DDPG+HER.

All of the tests show that AACHER surpasses the traditional DDPG+HER. AACHER tops all other algorithms based on success rate, reward, and average Q value.

\section{Conclusion and Future Work}\label{chapter_six}
In this paper, we proposed a unique algorithm AACHER (Actor-Critic Deep Reinforcement Learning with Hindsight Experience Replay). Actor/critic learning is important for DDPG's performance. 
In our algorithm AACHER, we use numerous independent actors and critics for DDPG, which is helpful to reduce the effects of one actor or critic performing poorly. Additionally, we combined HER and our proposed DDPG in AACHER. The actor-critic learning architecture used by the AACHER approach is more dependable, which results in a more stable training environment and better performance in real-world situations. To evaluate the effectiveness of our algorithm, we conducted a variety of experiments. \textit{AuboReach}, \textit{FetchReach-v1}, \textit{FetchPush-v1}, \textit{FetchSlide-v1}, and \textit{FetchPickAndPlace-v1} were the five simulated environments in which we evaluated AACHER. Among all the instances made for DDPG, A10C10 and A20C20 had the best performance. The overall results showed that AACHER performs well in all three performance evaluation matrices and outclasses the traditional DDPG+HER in all situations.

Additionally, the AACHER approach is simple to integrate with several RL state-action value-related methods. It is understood that some of the hyper-parameters in AACHER might be problematic. Therefore, in future work, we will concentrate on making the loss function's parameters trainable variables. In addition, we will explore HER to find a better experience replay mechanism.

\bibliographystyle{ACM-Reference-Format} 
\bibliography{sample}


\end{document}